%% file: bare_jrnl_compsoc.tex
\documentclass[10pt,journal,compsoc]{IEEEtran}
\usepackage{cite}

\ifCLASSINFOpdf
   \usepackage[pdftex]{graphicx}
   \graphicspath{{fig_eps/}{eps_img/}{add_img/}{author_img/}}
\else
   \usepackage[dvips]{graphicx}
   \graphicspath{{fig_eps/}{eps_img/}{add_img/}}
   \DeclareGraphicsExtensions{.eps}
\fi

\usepackage{amsmath}

\usepackage{bm}
\usepackage{}
\usepackage{fixltx2e}

\usepackage{stfloats}

\usepackage{caption}

\usepackage{subfigure}

\usepackage{graphicx}

\begin{document}
%
\title{Deep3DPose: Realtime Reconstruction of Arbitrarily Posed Human Bodies from Single RGB Images}
%
%
%
%

\author{Liguo~Jiang,
		Miaopeng~Li,
		Jianjie~Zhang,
		Congyi~Wang,
		Juntao~Ye,
		Xinguo~Liu,
		and~Jinxiang~Chai
\IEEEcompsocitemizethanks{
	\IEEEcompsocthanksitem L. Jiang, J. Ye are with NLPR, Institute of Automation, Chinese Academy of Sciences, Beijing, China and School of Artificial Intelligence, University of Chinese Academy of Sciences, Beijing, China. \protect\\ 
	E-mail: jiangliguo2015@ia.ac.cn and yejuntao@gmail.com.
\IEEEcompsocthanksitem J. Zhang and C. Wang are with Xmov, Shanghai, China.
\IEEEcompsocthanksitem M. Li and X. Liu are with State Key Laboratory of CAD\&CG, Zhejiang University, Hangzhou, China.
\IEEEcompsocthanksitem J. Chai is with Texas A\&M University.
}
\thanks{}}

%
%

\markboth{IEEE TRANSACTIONS ON VISUALIZATION AND COMPUTER GRAPHICS,~Vol.~XX, No.~, August~2020}%
{Jiang \MakeLowercase{\textit{et al.}}: Deep3DPose: Realtime Reconstruction of Arbitrarily Posed Human Bodies from Single RGB Images}

%



\IEEEtitleabstractindextext{%
\begin{abstract}
We introduce an approach that accurately reconstructs 3D human poses and detailed 3D full-body geometric models from single images in realtime. The key idea of our approach is a novel end-to-end multi-task deep learning framework that uses single images to predict five outputs simultaneously: foreground segmentation mask, 2D joints positions, semantic body partitions, 3D part orientations and $uv$ coordinates ($uv$ map). The multi-task network architecture not only generates more visual cues for reconstruction, but also makes each individual prediction more accurate. The CNN regressor is further combined with an optimization based algorithm for accurate kinematic pose reconstruction and full-body shape modeling. We show that the realtime reconstruction reaches accurate fitting that has not been seen before, especially for wild images. We demonstrate the results of our realtime 3D pose and human body reconstruction system on various challenging in-the-wild videos. We show the system advances the frontier of 3D human body and pose reconstruction from single images by quantitative evaluations and comparisons with state-of-the-art methods.
\end{abstract}

}

\maketitle

\IEEEdisplaynontitleabstractindextext

%
\IEEEpeerreviewmaketitle

\input{1-Introduction}
\input{2-RelatedWork}
\input{3-Methods}
\input{4-Reconstruction}
\input{5-Results}

\ifCLASSOPTIONcaptionsoff
  \newpage
\fi



\bibliographystyle{IEEEtran}
\bibliography{IEEEabrv,./human_bibliography0515}

%

\begin{IEEEbiography}[{\includegraphics[width=1in,height=1.25in,clip,keepaspectratio]{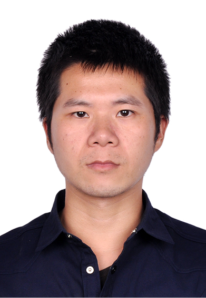}}]{Liguo Jiang}
received the B.Eng degree in software engineering from Chongqing University in 2015. He is currently working toward the PhD degree in National Laboratory of Pattern Recognition, Institute of Automation, Chinese Academy of Sciences. His research interests include deep learning, human motion capture and cloth simulation. 
\end{IEEEbiography}

\begin{IEEEbiography}[{\includegraphics[width=1in,height=1.25in,clip,keepaspectratio]{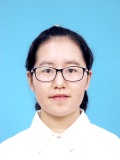}}]{Miaopeng Li}
	is a Ph.D. student at the State Key Lab of CAD\&CG, Zhejiang University, China. She received her bachelor degree from Northwestern Polytechnical University in 2016. Her research interests include marker-less human motion capture, human pose estimation, 3D reconstruction and their applications.
\end{IEEEbiography}

\begin{IEEEbiography}[{\includegraphics[width=1in,height=1.25in,clip,keepaspectratio]{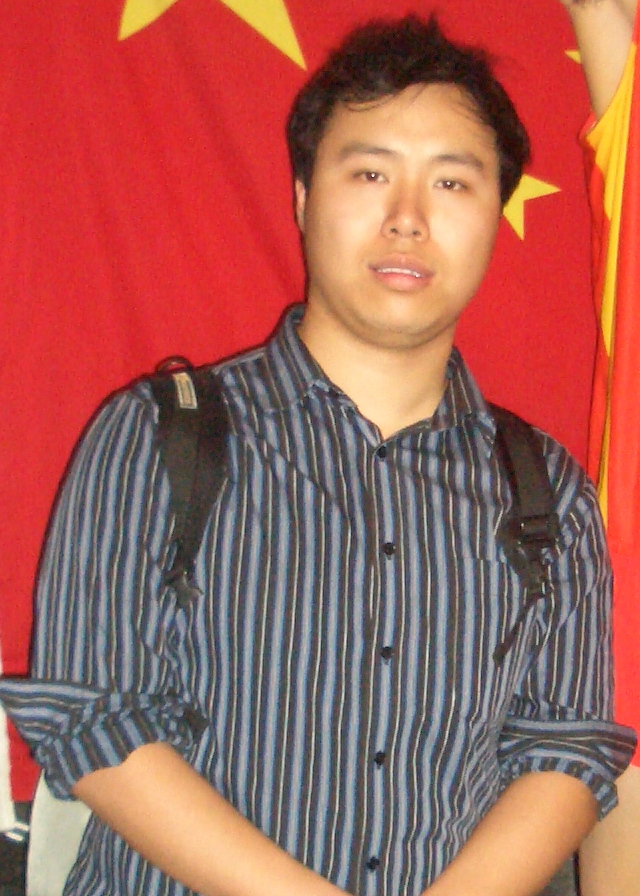}}]{Jianjie Zhang} received PhD degree in computer science from Texas A\&M Univeristy (TAMU). He is currently a R\&D director in Xmov ai Inc. His primary research is in the area of computer graphics and vision, including human body modeling and tracking, human body dynamics simulation, human face modeling and tracking and etc.
\end{IEEEbiography}

\begin{IEEEbiography}[{\includegraphics[width=1in, height=1.25in,clip,keepaspectratio]{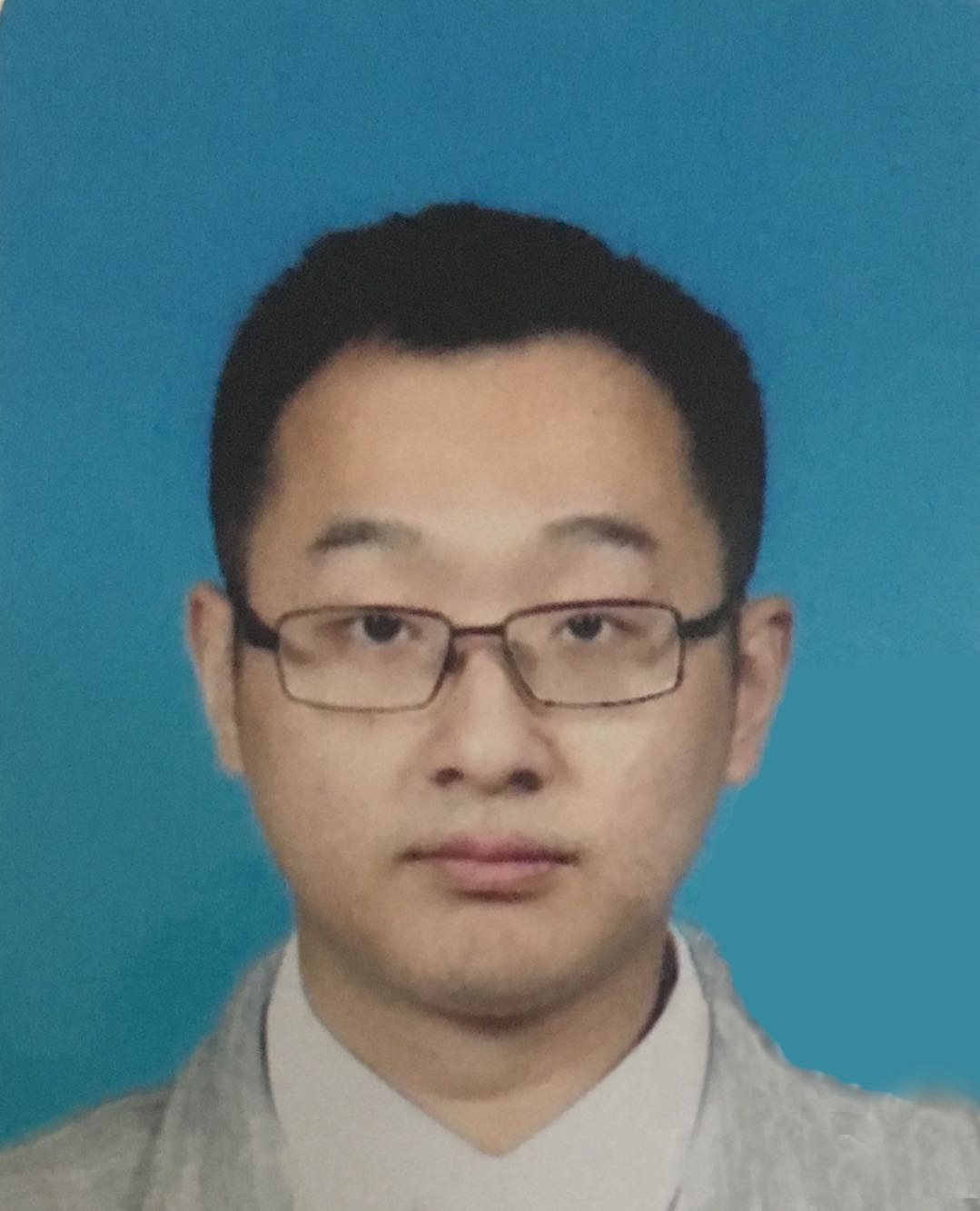}}]{Congyi Wang}
	received the PhD degree in computer science from Institute of Computing Technology, Chinese Academy of Sciences in Jan 2017. Since 2018, he has been a research scientist at XMov, a startup company aiming at AI powered virtual production line. His research interests include computer animation, computer graphics, computer vision and speech signal processing.
\end{IEEEbiography}

\begin{IEEEbiography}[{\includegraphics[width=1in,height=1.25in,clip,keepaspectratio]{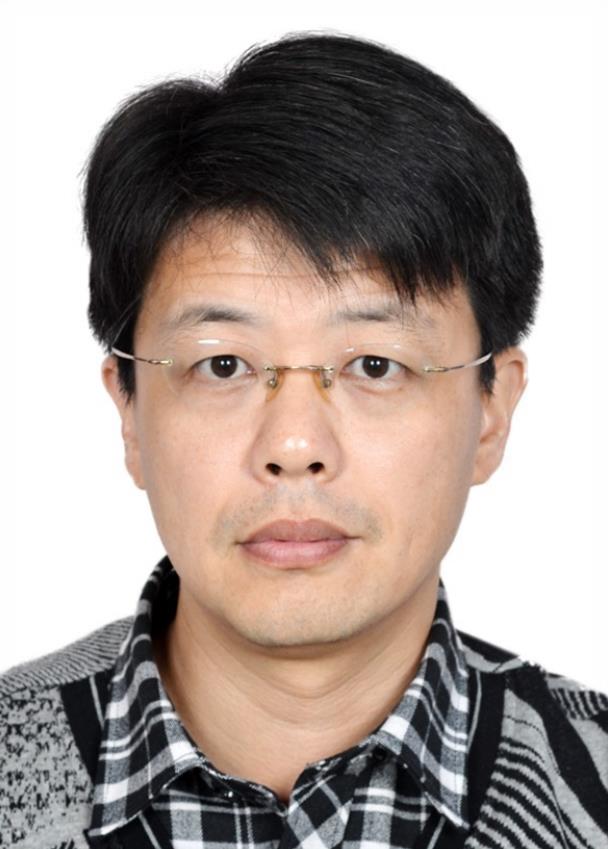}}]{Juntao Ye}
	was awarded his B.Eng from Harbin Engineering University in 1994, MSc from Institute of Computational Mathematics and Scientific/Engineering Computing, Chinese Academy of Sciences in 2000, and his PhD in Computer Science from The University of Western Ontario, Canada, in 2005. He is currently an associate professor with National Laboratory of Pattern Recognition of the Institute of Automation, Chinese Academy of Sciences. His research interests include graphics, particularly physicallybased simulation of cloth and fluid, as well as image/video processing.
\end{IEEEbiography}

\begin{IEEEbiography}[{\includegraphics[width=1in,height=1.25in,clip,keepaspectratio]{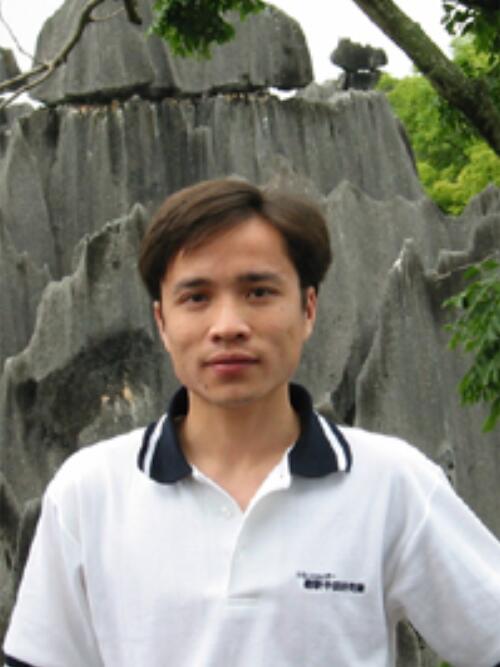}}]{Xinguo Liu}
	received the BS and PhD degrees in applied mathematics from Zhejiang University in 1995 and 2001, respectively. He is a professor at the School of Computer Science and Technology, Zhejiang University. He was with Microsoft Research Asia in Beijing during 2001-2006, and then joined in Zhejiang University. His main research interests are in graphics and vision, particularly geometry processing, realistic and image-based rendering, and 3D reconstruction.
\end{IEEEbiography}

\begin{IEEEbiography}[{\includegraphics[width=1in,height=1.25in,clip,keepaspectratio]{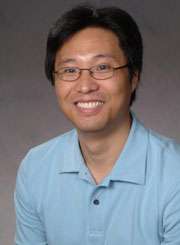}}]{Jinxiang Chai}
	received PhD degree in computer science from
	Carnegie Mellon University(CMU). He is currently an associate
	professor in the Department of Computer Science and Engineering
	at Texas A\&M University. His primary research is in the area of
	computer graphics and vision with broad applications in other
	disciplines such as virtual and augmented reality, robotics,
	human computer interaction, and biomechanics. He received
	an NSF CAREER award for his work on theory and practice of
	Bayesian motion synthesis.
\end{IEEEbiography}

%
%






\end{document}

%% file: 1-Introduction.tex
\IEEEraisesectionheading{\section{Introduction}\label{sec:introduction}}
\IEEEPARstart{C}{reating} natural-looking human characters with realistic motions is essential for many applications, including movies, video games, robotics, sports training, medical analytics and social behavior recognition, and so on.
Using expensive and special equipment, such as multi-cameras and reflective markers based motion capture systems, this task can be achieved without too much pain for scenes that do not impose many restrictions.
Yet the inconvenient accessibility to such equipment has limited the flourishing of 3D human motion related applications.

The ideal and most convenient way is to use off-the-shelf RGB cameras to capture live performance and create 3D motion data.
The minimal requirement of a single RGB camera is particularly appealing, as it offers the
lowest cost, easy setup, and the potential of converting huge volume of Internet videos into a large-scale 3D human body corpus.
Recent years have seen much research efforts being devoted to estimating not only the skeletal motion but also body pose and shape.
Yet reconstructing 3D pose and shape from a single RGB camera is still a challenging and underconstrained problem with inherent ambiguities, 
especially in wild uncontrolled environment and in realtime.
Therefore the state-of-the-art results are often vulnerable to ambiguities in the video (e.g., occlusions, cloth deformation,
and illumination changes), degeneracy in camera motion, and a lack of discernible features on a human body.
Moreover, methods that achieve realtime, robust as well as accurate performance have rarely been seen common so far.

We introduce an approach that is capable of obtaining accurate 3D human poses and body shape from single wild images in realtime.
When applied to video sequences, our system outputs temporally consistent bodies in motion at more than 20 Hz on a desktop computer.
The power of our method comes from a convolutional neural network (CNN) which leverages a multi-task architecture that is able to outputs five results simultaneously:
foreground mask, 2D joint positions, body partition, 3D part orientation fields (POFs) and $uv$ coordinates.
Body partition index and $uv$ coordinates indicate part-specific $uv$ coordinates, which is called IUV \cite{guler2018densepose}.
While none of existing networks support so many tasks at one time, this architecture makes it possible to refine multiple predictions recurrently.
The regressed results are fed into a kinematic skeleton pose and body geometry fitting optimizer and outputs a camera-relative full 3D posed body mesh.
The success of our approach also relies on the expansion of publicly available training datasets.
While it is feasible to annotate a small number of labels in 2D images, upgrading to a large number of 3D representation becomes impractical.
The new data is collected with our in-house cost-efficient, marker-less and scalable data acquisition system, and is preprocessed efficiently.

\begin{figure*}
	\includegraphics[width=0.99\linewidth]{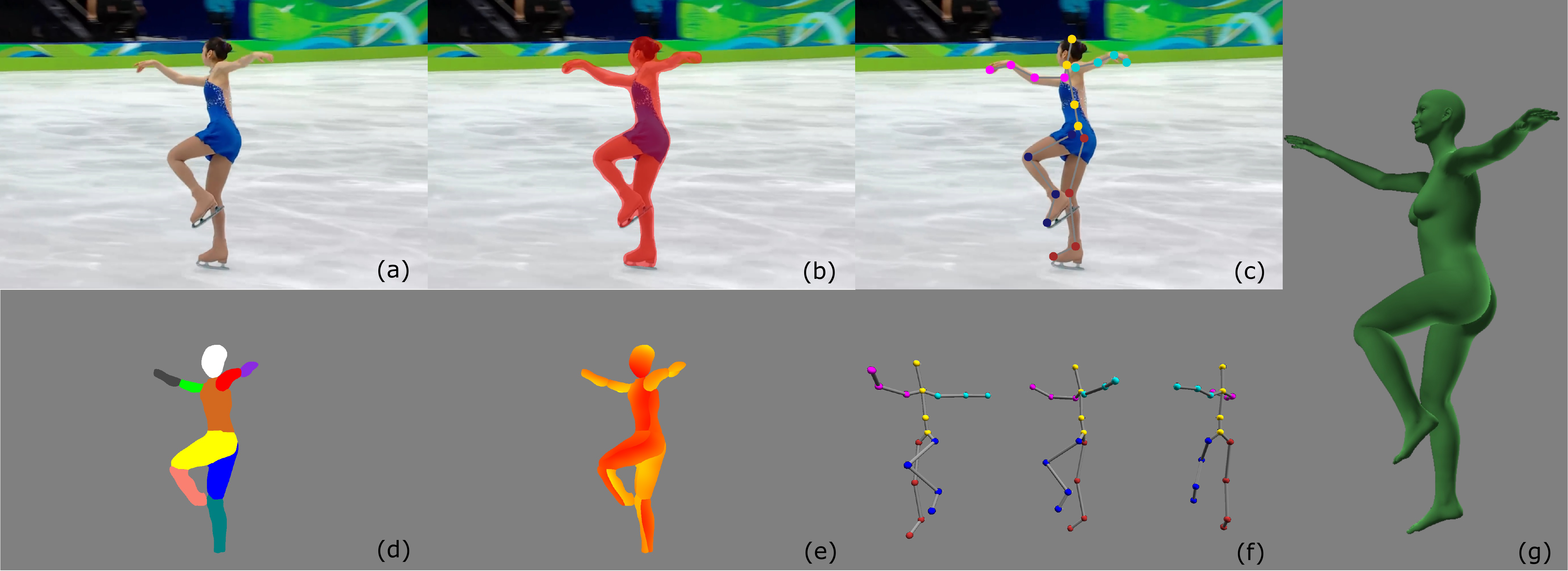}
	\caption{Given an image (a), our regression network produces five outputs simultaneously: foreground segmentation mask (b), 2D joints positions (c), body partition (d), and a $uv$ map (e), 3D part orientations (applied to a mean skeleton) (f). These outputs further guide the generation of a full-body model (g). The whole process runs in realtime on a desktop computer. }
	\label{fig_teaser}
\end{figure*}

The power of our system is demonstrated by reconstructing 3D human poses and shapes for a wide variety of subjects from monocular video sequences.
We have tested our realtime system on both live video streams and the Internet videos, demonstrating its accuracy and robustness
under a variety of uncontrolled illumination conditions and backgrounds, as well as significant variations on races, shapes, poses, clothes across individuals.
We show that our system can reconstruct bodies with realistic poses for highly dynamic motions such as figure skating (Fig.~\ref{fig_teaser}),
low energy motions such as walking, and motions with human-environment interaction such as sitting and standing up. We evaluate the importance of each key component of our algorithm, by dropping off each component in the reconstruction.
We advance the state-of-the-art realtime reconstruction of 3D human poses and detailed geometric body meshes from single images, and offer comparisons with alternative solutions \cite{kolotouros2019learning,mehta2017TOG,bogo2016ECCV,kocabas2020vibe}.

The highlights of our 3D reconstruction system are
\begin{itemize}
\item {\bfseries Realtime.}
Thanks to our specially designed neural network, we are able to regress multiple human structural features from single images in realtime.
We further feed network outputs to an efficient 3D human pose and body geometry fitting optimizer, and achieve realtime reconstruction performance.
\item {\bfseries Fully automatic and robust.}
With the abundant regression outputs per-frame, reconstruction can be achieved from one single image, independent of any pre-initialized state. 
This makes reconstruction from videos no longer suffers from the headache of re-initialization. Our system is also robust to illumination variation, as well as clothing diversity.

\item {\bfseries Accuracy.}
Our realtime system achieves reconstruction quality that is even more accurate than most offline or video-based methods in wild images.
This achievement is mainly due to three points: (1) a novel multi-task deep learning network
predicts abundant features, which boosts each other;
(2) an efficient reconstruction process that seamlessly integrates all the visual features obtained from the deep learning network.
(3) the augmentation to existing training dataset with our newly collected data.
\end{itemize}

%% file: 2-RelatedWork.tex
\section{Related Work}

\begin{figure*}
	\centering
	\includegraphics[width=1.9\columnwidth]{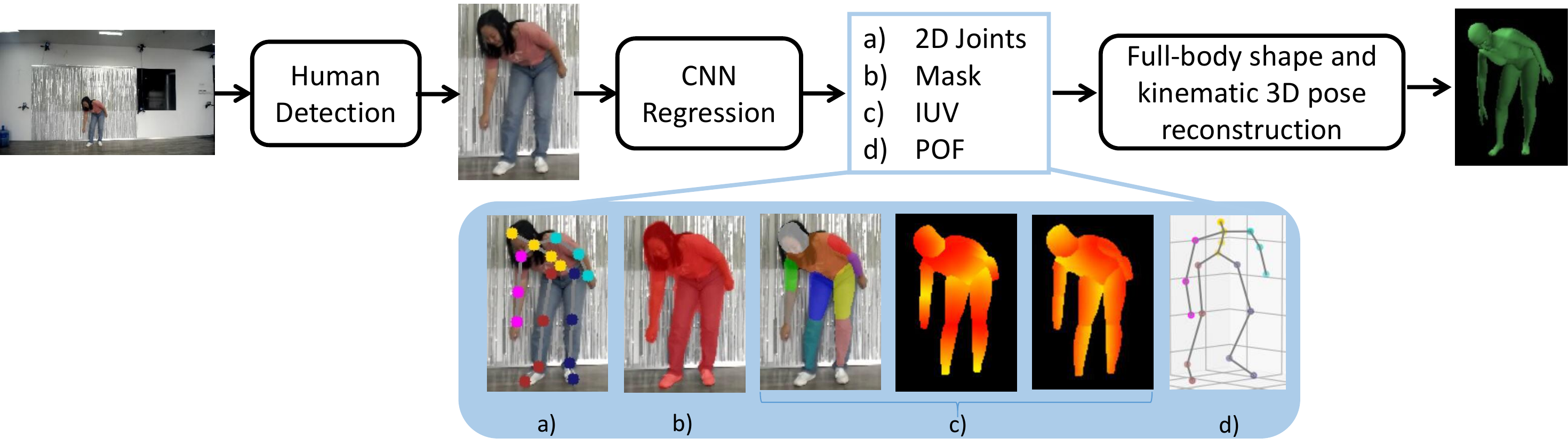}
	\caption{System overview. The CNN outputs 2D joints, foreground mask, IUV (including body partition and $u$-map and $v$-map) and POF (Part Orientation Field). }
	\label{fig:pipeline}
\end{figure*}

The research on human body reconstruction from single RGB images is traced back to skeleton joints estimation, from 2D to 3D,
and has achieved significant advances in recent years.
This line of work has further boosted the interest for simultaneous pose and shape estimation.
We will focus our review on 2D pose estimation, 3D pose and body reconstruction from single images.

\textbf{2D Pose Estimation.}
Nowadays image based single-person estimation \cite{wei2016CVPR,newell2016ECCV,nie2018human} has achieved great improvement due to the success of CNN.
These methods usually regress a probability map for each joint, designating the probability of a joint being located on each image pixel.
The image-to-surface correspondence (IUV), represented by human part partition and $uv$ coordinates map, was proposed in Densepose \cite{guler2018densepose}, and it is more effective and expressive than positioning just sparse 2D joints.
By predicting the $(u,v)$ coordinates and body part index for each pixel, a dense correspondence between pixels and points on a 3D mesh is defined. 
Our goal is different from these 2D or dense pose estimation methods in that we focus on 3D pose and geometry model reconstruction.

\textbf{3D Pose estimation.}
Other than regressing 2D pose or dense pose from single images, many people attempt to estimate 3D pose directly from images.
Most recent works can be divided into two categories: the {\itshape one-stage method} and the {\itshape two-stage method}.
In the two-stage methods \cite{martinez2017ICCV,ci2019optimizing, zhao2019semantic,fang2018learning}, the task of 3D pose estimation is decoupled into 2D joint detection and 3D coordinate regression. However, due to ambiguity of 3D estimation from 2D joints, these methods not only overlook certain image features having 3D cues, but also are very sensitive to the results of 2D pose estimation.
To overcome the ambiguity in lifting 2D to 3D, priors are introduced in some works.
Pavlakos et al. \cite{pavlakos2018CVPR} further annotated the ordinal depth relation in the COCO dataset \cite{502} and the MPII dataset \cite{andriluka14cvpr},
and proposed to estimate not only the 2D pose but also the ordinal depth relation as the extra input for lifting 2D to 3D. They achieved much better results.
Different from \cite{pavlakos2018CVPR}, joint limits and bone lengths are introduced as constraints in \cite{zhou2017towards}.
The one-stage methods usually use a single cropped image as the input to a CNN, and directly obtain root-relative 3D joint positions \cite{mehta20173DV, tekin230311, habibie2019in}, parent-relative joint positions \cite{sun2017ICCV}, or voxel joint probability map \cite{pavlakos2017CVPR, sun2018ECCV}.
In the VNect method \cite{mehta2017TOG}, a fully convolutional network structure is proposed to directly regress location maps, in order to decrease the dependency on tight bounding boxes for human. More importantly, VNect gets global coordinates rather than root-relative coordinates, and produces real-time performance.
The OriNet \cite{luo2018orinet} decouples bone lengths and bone orientations by representing 3D poses with 3D orientations of limbs, which are very suitable for motion control. We also adopt the representation of 3D orientations of limbs. Yet different from above 3D pose estimation network, we design an end-to-end network to regress a foreground mask, 2D joint positions, body partition, $uv$ coordinates and 3D part orientations simultaneously. Please note that existing works address only one or a subset of the tasks that we address. Multi-task learning usually boosts the quality of each individual output due to the correlation among tasks, and our experiments witness this fact.
On the other hand, while their goal focuses on 3D pose estimation only, we further reconstruct human body geometry automatically. With these image features and the geometry model, we are able to obtain much more accurate pose reconstruction with strong temporal fitting.

\textbf{Model-based Pose Estimation.}
Our method is related to one set of {\itshape model-based pose estimation} methods. 
Such approaches consider a parametric model of the human body, like SCAPE \cite{Anguelov2005SCAPE}, SMPL \cite{loper2015TOG} and SMPL-X \cite{SMPL-X:2019}, and the goal is to reconstruct a full 3D body pose and shape.
These approaches can be further divided into {\itshape model-based optimization} and {\itshape model-based regression}.

In the first category, \cite{Guan2009iccv} relies on annotated 2D ground truth, including joint landmarks and body silhouettes, to optimize the parameters of the SCAPE model through minimizing errors of the reprojected evidence.
With the SMPLify approach \cite{bogo2016ECCV}, this procedure was made automatic by replacing annotated 2D joints with 2D pose estimator. The whole process is then independent of user interference. Moreover, inter-penetration constraints are introduced to decrease the depth ambiguity when lifting 2D joints to 3D. The human shape estimation in SMPLify, however, relies on 2D joints only and does not constrain the body shape completely.
To address this issue, UP3D \cite{Lassner2017cvpr} further extends the SMPLify method by adding human silhouette to estimate human shape parameters, with the pipeline being still automatic.
Because of the binary representation of the human silhouette, as well as the introduction of cloth intervention in this method, the body shape is still not sufficiently constrained. To overcome these issues, two mechanisms have been introduced by our method. The first one is the IUV, which is albeit more expensive but provides a dense correspondence between an image and a 3D model. The second one is the 3D limb orientation, which makes the reconstruction of human shape and pose more precise.

Among the model-based regression methods, HMR \cite{kanazawa2018CVPR} uses a weakly supervised approach to regress the SMPL parameters directly from images, relying on 2D keypoints reprojection and a pose prior learnt in an adversarial manner.
Instead of regressing SMPL parameters directly, CMR \cite{kolotouros2019convolutional} builds a structure with Graph-CNN to model the connection of adjacent vertices of a human body mesh, and the 3D coordinates of mesh vertices are directly regressed.
EFT \cite{joo2020exemplar}, on the other hand, attempts to enrich wild images with missing SMPL parameters. 
By fine-tuning the HMR for each wild image, a few iterations to minimize errors of the 2D projection, and the current SMPL parameters are obtained. Treating these parameters as the ground truth for wild images, the original HMR is fine-tuned for the whole wild datasets. 
SPIN \cite{kolotouros2019learning} adopts a similar idea. Rather than fine-tuning the HMR to get the ground truth for wild images, SPIN use the optimization-based method, like SMPLify \cite{bogo2016ECCV}, to refine the result to be used by HMR as ground truth. 
Instead of directly regressing highly non-linear shape and pose parameters from an image, we regress multiple image features, and get body shape and pose by a well-designed optimization formulation. The experiments show that our reconstruction results are much more accurate, and also stable on image sequences.

\textbf{Model-based tracking.} 
Our work is also related to model-based tracking of 3D human poses using a single RGB camera. 
Usually this type of method pre-defines a human skeleton/body on initialization, and the 3D pose is updated by minimizing the inconsistency between the hypothesized poses and observed 2D measurements \cite{bregler2004twist}. This method, on one hand, needs a careful initialization; on the other hand the optimization is prone to get stuck in local minima, leading to track failures in the coming frames.
What is different in our method is that the body model is reconstructed automatically, and is not sensitive to initialization under abundant constraints. Accurate results are obtained from single images, therefore it is also robust on image sequences.

Several other works combine the power of regression and fitting, as we do.
The Total Capture method \cite{xiang2019CVPR} regresses 2D joint positions and 3D limb orientations with a network, and then optimize human face, body, and hands with a unified model Adam \cite{kingma2014adam}. Our method outputs more predictions in realtime (e.g. the IUV and foreground mask), thus gives more accurate reconstruction of body and pose. More importantly, the goal of our system design is realtime, therefore we want to discuss more about realtime systems here.
VNect \cite{mehta2017TOG} is the first system that captures kinematic skeleton using a single RGB camera. It uses a fully convolutional network to regress 2D pose and 3D root-relative joint positions, and then fit for a kinematic skeleton. PhysCap \cite{PhysCapTOG2020} adds environment constraints for
VNect to ensure the biophysical plausibility of human postures. In contrast, our multi-task network outputs more features thus more expressive geometry models can be reconstructed, which is manifested by the experimental results. Based on human skeleton pose capture, some methods further capture non-rigid deformation of clothing using optimization such as MonoPerfCap \cite{xu2018TOG} and LiveCap \cite{Habermann2019LRH}, or regression such as DeepCap \cite{habermann2020deepcap:}. 
However, the dependency on a pre-scanned, pre-reconstructed and pre-rigged subject-specific model limits the usability of these methods, while our system is fully automatic.

\begin{figure*}[ht]
	\centering
	\includegraphics[width=1.8\columnwidth]{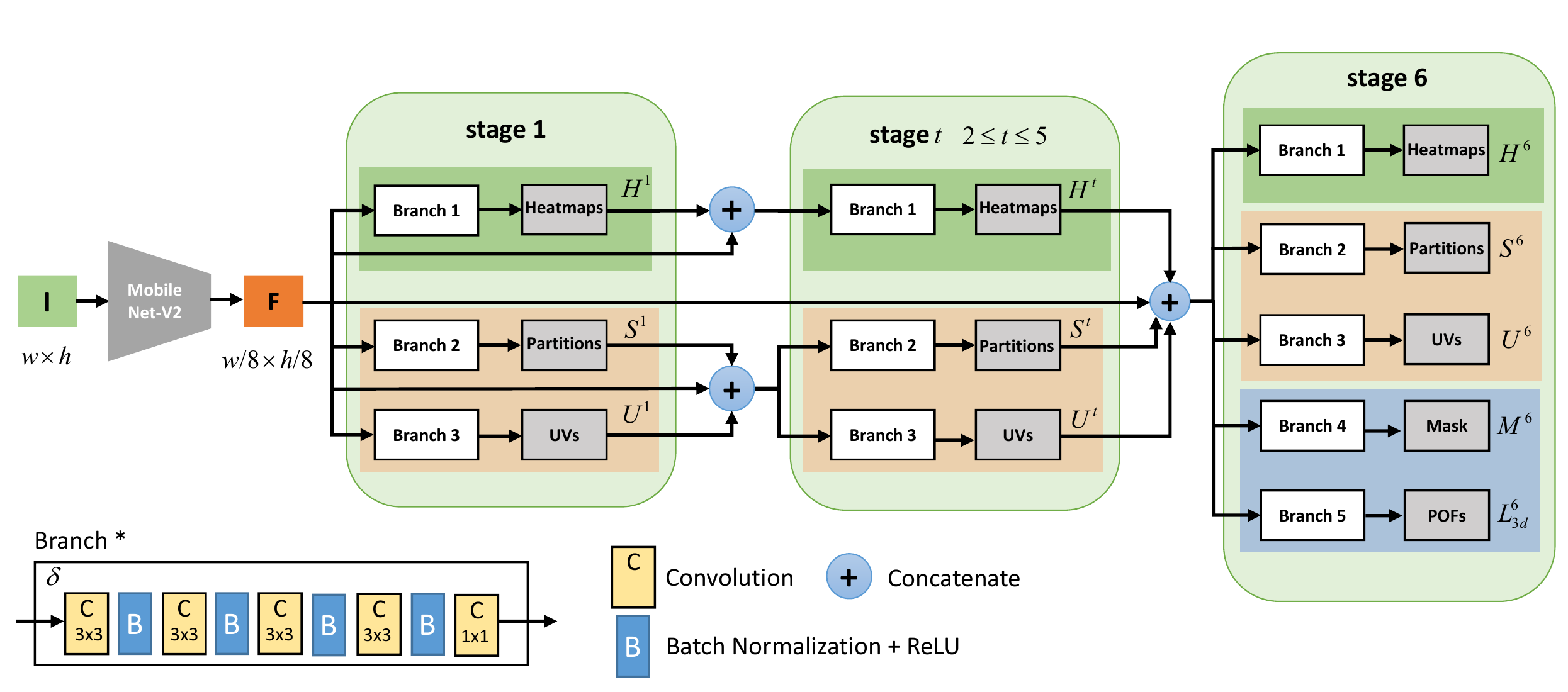}
	\caption{Architecture of our multi-task network.}
	\label{fig:network}
\end{figure*}

%% file: 3-Methods.tex
\section{Method Overview}

Our method takes as input a single RGB image with a single person, and outputs a 3D human body whose pose and shape are in accordance with the person in image.
This method consists of two parts: a neural network that regress measurements of human anatomical structures, and an optimization model that utilizes the network outputs to build a 3D body mesh.
For the regression part (\S~\ref{sec:network}), a human is first detected from image by YOLO \cite{redmon2016CVPR}, outputting a bounding box.
Then the cropped image is fed into a convolutional neural network (CNN) to get five outputs, namely foreground segmentation mask, 2D joints positions,
body partition, $uv$ coordinates, and 3D part orientations (which is also encoded as part orientation field \cite{luo2018orinet,xiang2019CVPR}). 
For the optimization part (\S~\ref{sec:reconstruction}), our method reconstructs body pose and shape by fitting a deformable human model.
We show that with as many as five features being integrated into the optimization pipeline,
the reconstruction reaches an accurate fitting that has not been seen before.
The whole pipeline is illustrated in Fig.~\ref{fig:pipeline}.
The success of our approach also relies on the enlargement of publicly available training datasets.
We describe how the new data is collected and preprocessed with our in-house acquisition system in \S~\ref{sec:training}.

\section{The Training Network}
\label{sec:network}

As mentioned, the key of our method is a multi-task CNN regressor for predicting five human anatomical structures:
foreground segmentation mask, 2D joints positions, body partition, $uv$ coordinates and 3D part orientations.
The motivation behind this multi-task architecture is that more outputs gives more visual cues to be used for reconstruction.
Actually this architecture refines multiple predictions recurrently, as a result each individual prediction turns to be more accurate.
This is not a surprise, as the power of multi-task learning is that efficiency and prediction accuracy can be improved
by learning multiple objectives from a shared representation \cite{caruana1998multitask}.

Our multi-task regressor is a fully convolutional network.
More specificlly, given a RGB image ${I \in {R}^{3 \times w\times h}}$, a feed-forward network simultaneously predicts a set of 2D joint confidence maps ${H \in {R}^{J \times w \times h}}$ (where $J=18$ is the number of joints to predict), human mask probability map ${M \in {R}^{w \times h}}$, human part probability map plus $uv$ map, and 3D Part Orientation Fields ${L \in {R}^{3O \times w \times h}}$, where $O=17$ is the number of body parts.

We use the term {\itshape IUV map} to indicate human partition probability map ${S \in {R}^{(C+1) \times w \times h}}$ (for $C=24$ partitions and one background) and $uv$ coordinates ${U \in {R}^{2C \times w \times h}}$, as did in \cite{guler2018densepose}.
To our knowledge, no previous works have ever regressed so many outputs as we do.

\subsection{Multi-task CNN Regression}
Fig.~\ref{fig:network} illustrates the structure of our multi-task network.
It is inspired by architectures like \cite{cao2017CVPR, wei2016pose, popa2017CVPR}, which refine the predictions recurrently.
An image is first encoded by a convolutional network, generating a set of image features $F$, which are then passed over to the first estimation for each individual task at stage 2. We get coarse predictions for joint confidence maps ${H^1}$, IUV maps (body partition ${S^1}$ and $uv$ coordinates ${U^1}$) in stage 1. In the successive stages, the network takes as input the image feature $F$, the results of previous stages of the same type. We formulates the procedure as follows:

\begin{equation}
H^t = \delta_{H}^{t}(Cat(F, H^{t-1}))
\end{equation}
\begin{equation}
S^t = \delta_{S}^{t}(Cat(F, S^{t-1}, U^{t-1}))
\end{equation}
\begin{equation}
U^t = \delta_{U}^{t}(Cat(F,S^{t-1}, U^{t-1}))
\end{equation}
where $2 \leq t \leq 5$ is the stage index, and $\delta(\cdot)$ is the mapping for Branch {*}, as defined as four Conv3$\times$3-BN-ReLU blocks and
one Conv1$\times$1 task-specified regressor. $Cat(\cdot)$ is the concatenation operation. In stage 6, the joint confidence map and IUV map from the previous stage
is concatenated and treated as input to predict not only the joint and IUV, but also two additional terms: the mask $M$ and the part orientation maps $L_{3d}$.

\textbf{Loss Term.} To guide the training of the multi-task network, we apply losses for predictions at each stage, specifically $L_2$ losses for the confidence maps $H$, POFs and $UV$ maps. Note for the $UV$ map, we only take into account a body part if the pixel is located inside it.
When training part partition, a standard multi-class cross-entropy loss is used. Note that due to the difference of human part areas, we balance the supervision for part segmentation classification by the weight ${w_c}$ for each human part $c$, so that the network would not over-fit body parts of large area. The balance weight ${w_c}$ is inversely proportional to the part area, as in \cite{yao2019densebody}. 
Our IUV (body partition and $UV$ maps) ground truth for hands is inaccurate, because we fit a statistic model into a skeleton without finger joints, so we just ignore the IUV loss for hands. The segmentation mask is trained by binary cross-entropy loss.

\begin{figure}
  \centering
  \includegraphics[width=0.96\columnwidth]{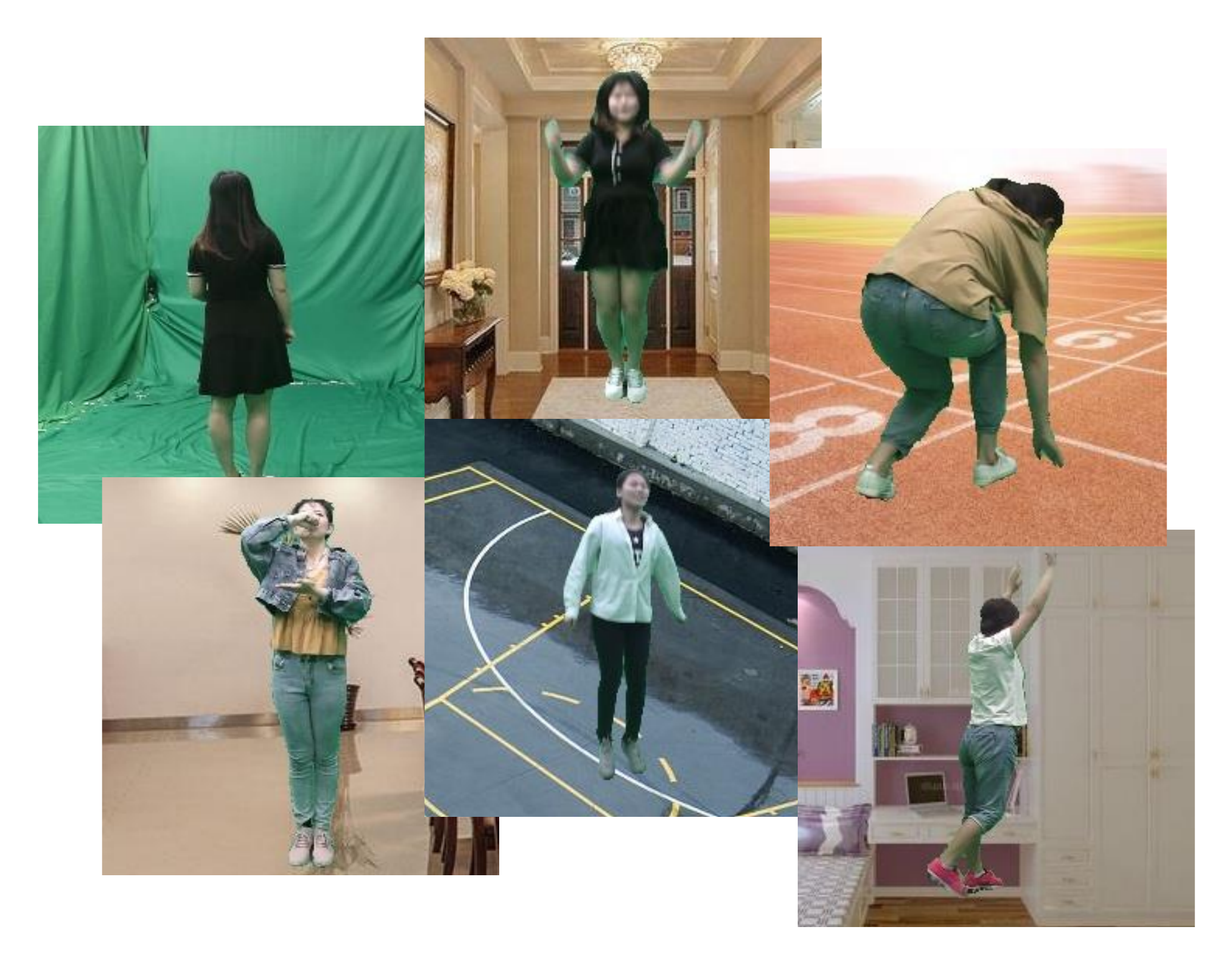}
  \caption{Our training dataset contains original captured images, as well as augmented images with background replacement.}
  \label{fig:seven}
\end{figure}

\textbf{Implementation.} The training of our multi-task network consists of three phases. (1) First, we pre-train our network for the 2D joint detection task with in-the-wild image dataset for stage $1\sim5$, ignoring other tasks, which gives better generalization performance. Our 2D joint detection task is trained with an initial learning rate of $10^{-3}$ and is reduced every 200,000 iters by a factor $\gamma=0.333$, as \cite{cao2018openpose} does.
(2) Second, we combine our newly collected dataset with 2D joint dataset, and apply a mix-training strategy for other tasks while freezing the weights of feature extractor and 2D joint detector for 100,000 iters by a learning rate of $5 \times 10^{-4}$. Note that our newly collected data has all desired ground truth for every task. Fig.~\ref{fig:seven} shows a few images in it, including the original captures and the augmented ones through background replacement. Data augmentation with background replacement greatly increases the generalization of in-the-wild images.
(3) Finally, we unfreeze the weights of well-trained 2D task and feature extractors, but apply a smaller learning rate (multiplied by 0.1) for these weights.
Our full-task training takes 1,000,000 iterations with a learning rate of $5\times10^{-4}$ reduced every 200,000 iters by a factor $\gamma=0.333$. 
We employ a rotation augmentation ($\pm 30^\circ$), a scaling augmentation (0.75-1.25) and left-right flipping (only for in-the-wild dataset) for training. 
We use the Caffe framework \cite{jia2014caffe} for network training, and use the Adadelta solver \cite{zeiler2012adadelta}. The performance of each task is strongly dependent on the relative weighting between the loss of each task \cite{kendall2018multi}. And in our experiment, we set loss weights as follows: $w=0.5$ for $UV$, $w=0.05$ for body partition, $w=0.5$ for heatmap, $w=1.0$ for foreground mask and $w=1.0$ for POFs. To balance accuracy and efficiency, we use MobileNet-V2 \cite{sandler2018mobilenetv2:} as our feature extractor.
Note that we removed the downsampling operations in last two blocks (by replacing stride=2 in downsampling convolution with stride=1), and maintained the size of the final feature map to be $1/8$ of the input image, which is 224-by-224.

%% file: 4-Reconstruction.tex
\section{Automatic Kinematic Pose Reconstruction and Full-body Shape Modeling}\label{sec:reconstruction}
Our work sets a new mark in terms of level-of-detail that previous work did not reach.
This mainly attributes to our innovative kinematic pose reconstruction and full-body shape modeling.
The network output (as in \S~\ref{sec:network}) is of low-quality and noisy, and may not be compatible with input images or with human kinematic constraints.
Directly using such information leads to inaccurate reconstruction of human motion.
To refine the network output, we develop a novel algorithm that accurately reconstructs the human motion as well as a subject-specific full-body mesh model.

\subsection{Human Full-body Representation}

Similar to SMPL, we approximate the human full-body geometry with a skinned mesh that is driven by an articulated skeleton model using Linear Blend Skinning (LBS).
Our skeleton has 45 degree of freedoms (DOFs), 6 of which are for global position and orientation, and 39 are for joint angles (note that a joint may be of 1, 2 or 3 DOFs).
We built a female mesh model of 28,109 vertices (or 56,142 triangular faces), carrying more geometric details than the SMPL model of $6,890$ vertices. The female model is elaborately rigged and parameterized such that the shape can be easily controlled. \footnote{Yet currently we do not have a decently parameterized male model on par with the female model. For scenes with a male subject, we either use motion-retargeting to drive a male model mesh but without body dimension adjustment (e.g. Fig.~\ref{fig:ablation_iuv}(b), or just blindly use the female model (e.g. one case in the accompanying video). }

\begin{figure}
	\centering
	\includegraphics[width=0.9\columnwidth]{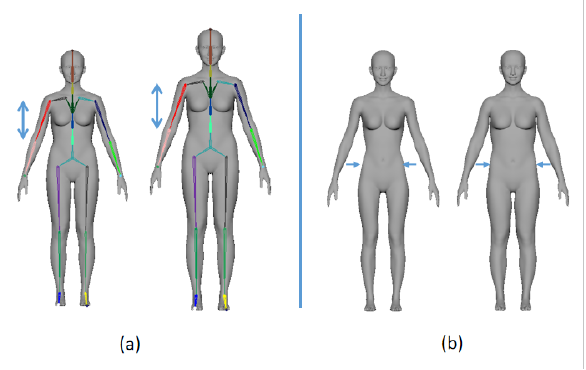}
	\caption{Human shape variations: (a) bone length variation; (b) body part thickness variation.}
	\label{fig:human_shape_model}
\end{figure}

Following a relatively mature process we build a parametric human full-body geometry model (Fig.~\ref{fig:human_shape_model}).
The model is controllable in two aspects: (1) skeleton scales, which encode coarse-level variation like the overall and the per-bone scales,
(2) mesh vertex offsets, which encode fine-level shape variation, such as thickness of a limb.
The parametric human full-body model can be represented as:
\begin{equation} \label{eq:HandModel}
\begin{split}
\textbf{H}_i({\bm \alpha}, {\bm \beta}, {\bm \theta}) =
\sum_{j=0}^{n-1} {
	w_{ij}T_j({\bm \theta}) \hat{\textbf{v}}_{ij}' }, \\
\hat{\textbf{v}}_{ij}' = \textbf{S}(\bm \alpha) \otimes (\textbf{Q}(\bm \beta) \oplus \hat{\textbf{v}}_{ij}),
\end{split}
\end{equation}
where $\textbf{H}_i(\cdot)$ is the coordinate of the $i$-th vertex of the mesh model, $\hat{\textbf{v}}_{ij}'$ is the result after applying the scaling and offsetting to $\hat{\textbf{v}}_{ij}$ (which is the $i$-th vertex represented in the local coordinate frame of the $j$-th bone),
$\textbf{Q}(\bm \beta) \oplus$ describes the vertex offsetting,
$\textbf{S}(\bm \alpha) \otimes$ describes the bone scaling,
the shape parameters ${\bm \alpha} \in R^8$ and ${\bm \beta} \in R^{26}$ provide a low-dimensional representation of human bone scale variances and vertex offset variances across individuals respectively,
${\bm \theta}$ is the pose for deformation, $T_j({\bm \theta})$ is the transformation of the $j$-th bone for pose ${\bm \theta}$,
$w_{ij}$ is a sparse weight map for deformation, $n$ is the number of bones.

\subsection{Kinematic Pose Reconstruction}\label{sec:KPR}

Given the subject-specific full-body mesh model (obtained in \S~\ref{sec:mesh_recon}) and the network observations (2D pose from 2D joints probability maps, 3D part orientations from the 3D limb orientation fields, the human mask, the IUV map for frame image $I_i$), our goal is to estimate a human pose ${\bm \theta}$ which best matches the network observations. We estimate the human pose ${\bm \theta}$ by minimizing the following objective function:
\begin{equation}
	\small
\mathop{\arg\min}_{\bm \theta} \ \ (w_{data} E_{data} \!+\! w_{prior} E_{prior} \!+\! w_{temporal} E_{temporal}).
\label{equation:kpr}
\end{equation}
where $E_{data}$ is the data term penalizing the registration error between the synthesized human model and the observation.
$E_{prior}$ is the prior term that penalizes invalid human pose configuration,
and $E_{temporal}$ is the pose smoothness term that penalizes the jerkiness in the motion, which is only used for video application.
While searching for the solution in an iterative manner, it is possible (and recommended) to use the pose ${\bm \theta} _{prev}$ from the previous frame as the initial guess. We will describe each term in detail in the following subsections.

\subsubsection{The Data Term}

The data term $E_{data}$ evaluates how well the current human pose ${\bm \theta}$ matches the network observations by the analysis-by-synthesis strategy.
Given the human pose ${\bm \theta}$, we first apply {\itshape skeleton subspace deformation} to synthesize a full-body mesh model.
And then we compute the registration error between the network observations and the synthesized human model. The data term is defined as
\begin{equation}
E_{data}({\bm \theta}) = w_{2d} E_{2d} \!+\! w_{3d} E_{3d} \!+\! w_{iuv} E_{iuv} \!+\! w_{mask} E_{mask} .
\label{eq:data_term}
\end{equation}
Here $E_{2d}$ and $E_{3d}$ are alignment constraints based on predicted 2D joints positions and 3D limb orientations, respectively. $E_{iuv}$ penalizes the registration error between the synthesized $uv$ map and the observed $uv$ map, and $E_{mask}$ penalizes error between the synthesized and the observed mask.

\textbf{Sparse 2D alignment.}
To minimize the discrepancy between estimated 2D joints $\hat{P}_{2d}$ and the projections of the 3D joints from the human body model, we incorporate $\hat{P}_{2d}$ into the following reprojection constraint:
\begin{equation}
E_{2d}({\bm \theta}) = \sum_i || \Pi (J_{3d}^i (\bm \theta)) - \hat{P}_{2d}^i ||_2^2 .
\end{equation}
Here $J_{3d}^i ({\bm \theta})$ is the $i$-th joint in the skeleton, and $\Pi$ is the 3D-to-2D projection matrix according to known intrinsic camera parameters.

\textbf{Sparse 3D alignment.}
Since many 3D poses share the same reprojection of 2D pose in single image, and it is hard to infer a 3D pose only with above mentioned reprojection constraint. Therefore we add another 3D constraint
\begin{equation}
E_{3d}({\bm \theta}) \!=\! \sum_{(m,n) \in B} || (||J_{3d}^m - J_{3d}^n||_{2} \cdot \hat{\mathbf O}_{m,n} - (J_{3d}^m - J_{3d}^n)) ||_{2}^2 .
\end{equation}
Here $\hat{\mathbf O}_{m,n}$ is the estimated 3D direction of limb $(m,n)$, which is the mean value along the segment from joint $J^m$ to $J^n$.

\textbf{Why do we use 3D direction?}
Various representations have been put forward to denote a 3D pose, including 3D joint positions, 2D joints position plus root-relative depth, and 3D limb directions and so on.
We adopt 3D limb direction due to its two advantages. First, limb orientation is scale-invariant and dataset independent, which helps resolve scale ambiguity and generalizes easily to diversity data. Second, because an auto-reconstruction of human shape is done before tracking,
there is no need to worry too much about limb length ratios in the following frames.

\textbf{Dense IUV alignment.}
In addition to the coarse level shape variations caused by bone length and pose change, another data term $E_{iuv}$ constrains fine-level shape variations,
such as the thickness of limbs. To this end, we construct a dense pixel-to-surface correspondence represented as an IUV map.
Each pixel of an IUV image has a body part index $i$, and a $(uv)$ coordinate that maps a pixel to a unique point on the surface of a body model.
Given an IUV map predicted by the neural network, we select some reliable pixels $p$ which satisfies $\delta^{(1)}(p) = 1$ for the function defined as
\begin{eqnarray}
\delta^{(1)}(p)\! =\!
\left\{
\begin{array}{ll}
1,   &  \mbox{if} \quad \mathtt{Max}\{{\mathbf v}(p)\} - \mathtt{Max2nd}\{{\mathbf v}(p)\} > \phi   \\
0,   &  \mbox{otherwise}    \\
\end{array} \right .
\end{eqnarray}
where ${\mathbf v}(p)$ is the segmentation probability vector of pixel $p$. In our experiment, we set threshold $\phi=0.5$.
Interpolating these image-to-surface points, a more accurate human model is obtained by minimizing
\begin{equation}
E_{iuv}({\bm \theta}) = \sum_p \delta^{(1)}(p) || \Pi (M({\bm \theta}, p)) - \hat{I}(p) ||_2^2 ,
\end{equation}
where $M({\bm \theta}, p)$ is the synthesized mesh vertex corresponding to pixel $p$, and $\Pi$ is the 3D-to-2D projection matrix according to known intrinsic camera parameters.

\textbf{Mask Term.}
The foreground segmentation mask term is to penalize the inconsistency between the mask of synthesized human mesh model and the mask from network observations.
\begin{equation}
E_{mask}({\bm \theta}) = \sum_v \delta^{(2)}(v) ||\Pi(v) - q_v||_2^2
\end{equation}
where $v$ is the mesh vertex, $\Pi$ is the 3D-to-2D projection matrix according to known intrinsic camera parameters, $\delta^{(2)}(v)$ indicating whether $v$ is outside the observed mask or not, $q_v$ is the corresponding 2D image position for $v$ obtained from the distance map of the observed mask.

\subsubsection{The Prior Term}
To make joints of a human skeleton to be physically meaningful, two different priors, the pose space prior and the joint limit, are defined and used to form the prior term
$$E_{prior} = w_{pose\_prior} E_{pose\_prior} + w_{jt\_limit} E_{jt\_limit} .$$

\textbf{Pose Space Prior.}
We construct individual PCA models for each body part (e.g., shoulders, arms, spines, legs and feet) via CMU mocap database\footnote{http://mocap.cs.cmu.edu/resources.php}. With those part-wise PCA models, we are able to constraint the solution space into the physically meaningful area by minimizing the following objective function:
\begin{equation}
E_{pose\_prior}({\bm \theta}) = ||P_k^T(P_k({\bm \theta} - {\bm \mu})) + {\bm \mu} - {\bm \theta}||_2^2,
\end{equation}
where ${\bm \mu}$ is the mean vector of the PCA model, and $P_k$ is the first $k$ principle components of the PCA model. Here $k$ is chosen to retain 95\% of original variations.

\textbf{Joint Limit.}
The joint limit term is added to penalize invalid joint poses that exceed the range of joint movement. Every joint angle $\theta _i$, $i=7,8...45$ should stay within $[\theta _i^l, \theta _i^u]$. The joint limit term $E_{jt\_limit}$ can be represented as
\begin{equation}
	\begin{split}
E_{jt\_limit}({\bm \theta}) = \sum _{i=7}^{45} (\delta^{(3)}(\theta _i < \theta _i^l) ||\theta _i - \theta _i^l||^2 \\
+ \delta^{(3)}(\theta _i > \theta _i^u) ||\theta _i - \theta _i^u||^2) ,
	\end{split}
\end{equation}

where the binary function $\delta^{(3)}(x)$ is 1 if $x$ is true, and is 0 if $x$ is false.

\subsubsection{Temporal Smoothness Term}
We add a smoothness term to penalize the pose jerkiness between two consecutive frames.
This smoothness term is defined as:
\begin{equation}
E_{temporal}({\bm \theta}) = ||{\bm \theta} - {\bm \theta} _{prev} ||_2^2
\end{equation}

\subsubsection{Optimization}
\label{sec:optimization}
As Eq.~\ref{equation:kpr} is represented as a sum of squares, we can efficiently solve it by Gauss-Newton method. Since every term is differentiable, we can directly compute the Jacobian matrix $J({\bm \theta})$ and then follow the standard Gauss-Newton step to solve $\delta{\bm \theta}$ and update current ${\bm \theta}$
\begin{equation}
\begin{aligned}
&J({\bm \theta})^T J({\bm \theta})\delta{\bm \theta} = J({\bm \theta})^Tr({\bm \theta}), \\
&{\bm \theta} = {\bm \theta} + \delta{\bm \theta},
\end{aligned}
\end{equation}
where $r({\bm \theta})$ is the residual vector formed by concatenating each term.

\textbf{Parameter values.}
In our implementation, the weights $w_{data}$, $w_{prior}$ and $w_{temporal}$ in Eq.~\ref{equation:kpr} are set to 1.0, 0.5 and 5.0 in our experiments.
Weight settings in Eq.~\ref{eq:data_term} are $w_{2d}=20.0$, $w_{3d}=300.0$, $w_{iuv}=2.0$ and $w_{mask}=1.0$. The weights in $E_{prior}$ are $w_{pose\_prior}=0.002$ and $w_{jt\_limit}=5.0$. The weight for temporal smoothness term is $w_{temporal}=0.003$.

\subsection{Full-body Shape Modeling}
\label{sec:mesh_recon}

Now we describe how to reconstruct a full-body mesh model for a subject using the network observations.
Note in the case that the input is a video stream, the shape reconstruction is done only once, for the first frame.

\subsubsection{Shape Reconstruction}
Given an image $I$, our goal is to reconstruct a subject-specific human body model $\textbf{H}({\bm \alpha}, {\bm \beta}, {\bm \theta})$, 
with the derived image positions for 2D joints, the segmentation mask, body partition, 3D part orientation,
and the dense correspondence ($uv$ map) for human parts.
We formulate the reconstruction as a non-linear optimization problem
\begin{equation}\label{equ:shaperecon}
\begin{split}
\mathop{\arg\min}_{{\bm \alpha}, {\bm \beta}, {\bm \theta}} \ \ (w_{data} E_{data} +  w_{pose\_prior} E_{pose\_prior} + \\  w_{shape\_prior} E_{shape\_prior}) ,
\end{split}
\end{equation}
where $E_{data}$ is the data term that describes the registration error between the synthesized model and the network observations that inherits from \S~\ref{sec:KPR}, $E_{pose\_prior}$ is the pose prior term that inherits from \S~\ref{sec:KPR}, $E_{shape\_prior}$ is the shape prior term that penalizes the deviation of shape parameters from the human shape in the database.

\textbf{The shape prior term.}
We model the shape prior distribution with multiple single-variable Gaussian models that is learnt from the human shape database.
We define $E_{shape\_prior}$ as the deviation distance with parameters
$$
E_{shape\_prior} = (\sum_i | \alpha ^i - \mu _{\alpha}^i |/\sigma_{\alpha}^i
                    + \sum_j | \beta ^j  - \mu _{\beta}^j | /\sigma_{\beta}^j) ,
$$
where $\mu$ is the mean value and $\sigma$ is the standard deviation.

\subsubsection{Optimization}
Directly optimizing Eq.~\ref{equ:shaperecon} is not efficient and often falls into local minima, because the pose and the shape are coupled. To address this issue, we decouple the optimization into two sub-optimization problems: pose optimization and shape optimization. In each iteration, we first fix shape parameters and optimize pose parameters, and then vice versa.

\textbf{Pose estimation.}
In this step, we optimize the pose parameter ${\bm \theta}$ while fixing the shape parameter ${\bm \alpha}, {\bm \beta}$. This process is identical to that in \S~\ref{sec:KPR}.

\textbf{Shape estimation.}
In this step, we optimize the shape parameter ${\bm \alpha}, {\bm \beta}$ while keeping the pose parameter ${\bm \theta}$ fixed. And therefore the optimization problem can be represented by
\begin{equation}\label{equ:sr}
\mathop{\arg\min}_{{\bm \alpha}, {\bm \beta}} (w_{data} E_{data} + w_{shape\_prior} E_{shape\_prior}) .
\end{equation}
Following the optimization method in \S~\ref{sec:optimization}, we can solve the shape parameters ${\bm \alpha}$ and ${\bm \beta}$.

\textbf{Parameter values.}
For shape reconstruction, the energy terms $E_{data}$ and $E_{pose\_prior}$ are the same as in \S~\ref{sec:KPR}. The shape prior weight we use is
$w_{shape\_prior}=0.1$.

\section{Training Data}
\label{sec:training}

As one major highlight of our work, we complement existing datasets by building a dataset with a large number of actors, everyday clothing appearances, a broad range of motions.
The data capture setup eases the appearance augmentation and extends the captured variability.
This gives a potential to bring significant boost to accuracy and generalizability of the learnt models.
In this section we describe the capture environment and the recording process, as well as the processing of the captured data.

\subsection{Experimental Setup}

\begin{figure}
	\centering
	\includegraphics[width=0.88\columnwidth]{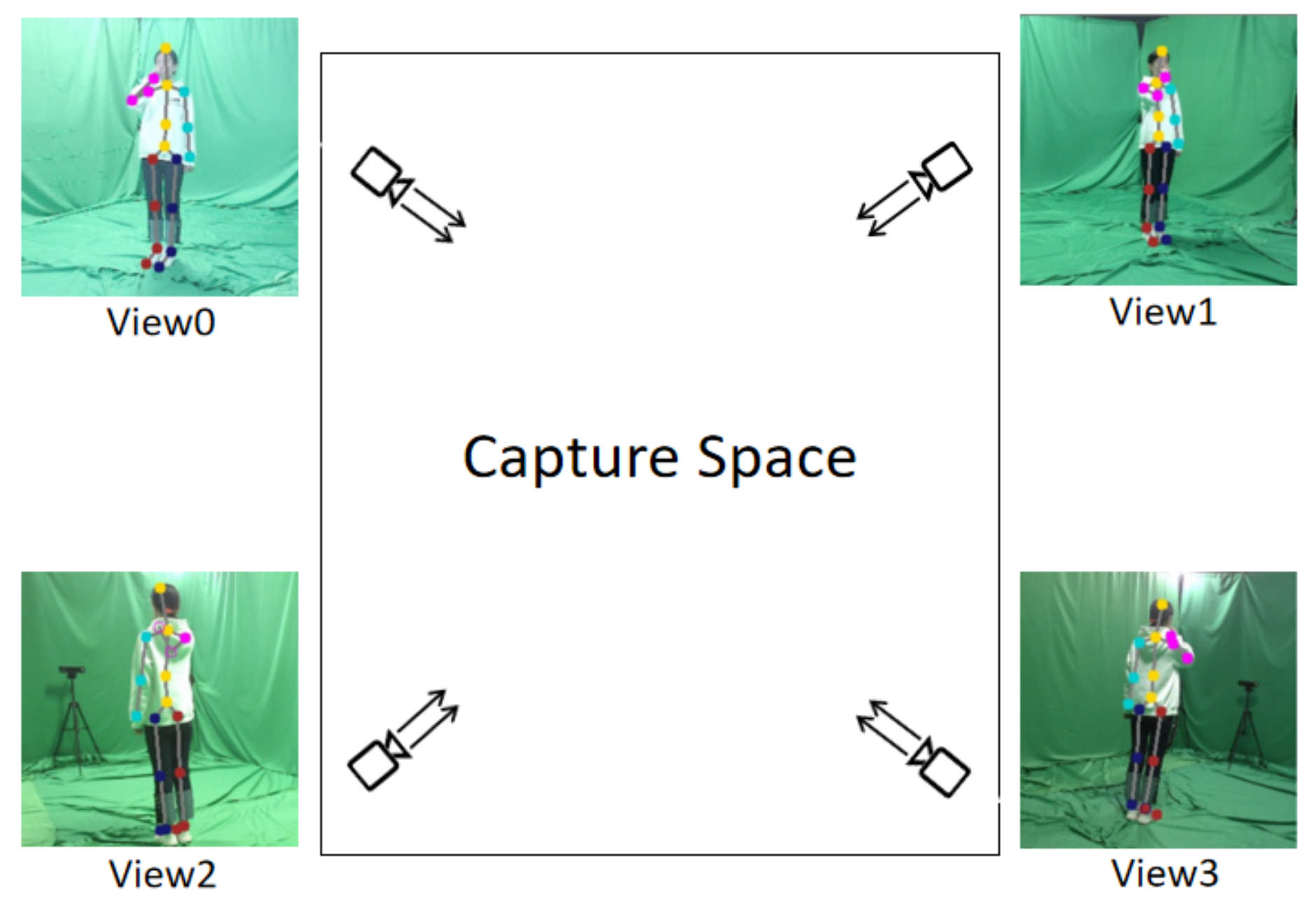}
	\caption{The capture space.}
	\label{fig:one}
\end{figure}

Our laboratory setup is shown in Fig.~\ref{fig:one}, where data is captured by four digital video cameras.
The designated laboratory area is about $4m \times 4m$, and within it we obtain effective capture court of approximately $2.5m \times 2.5m$,
where each subject is fully visible to all cameras.
Four cameras are placed at four corners of the court.
The floor and the wall are mantled with green curtains, making it easy for automatic segmentation of the foreground body.
The total number of actors screened are over $300$, covering a broad range of ages, body shapes and pose extensions.
Our dataset has much more subjects than in any of existing 3D datasets. Each person is asked to do certain daily life motions as well as sport motions,
for about three minutes.
To eliminate redundancy between consecutive video frames, frame images are further filtered, and only 500 to 1,000 images are sampled for each actor.
The sampling is achieved by clustering frame images according to the similarity of human poses.

\subsection{Shape Reconstruction from Measurements}

To build a highly accurate parametric body model for each actor, we take some body measurements while an actor is standing still in A-pose.
A set of measurements $\mathbf{M} \in R^{44}$ (including but not limited to lengths of limbs, shoulder and back, girths of chest, wrist, hip and stomach, etc.), are tailoring measured with a ruler.
Our parametric model has $8+26=34$ shape parameters, more sophisticated than the SMPL model,
which has only 10 shape parameters.
With the measurements, parameters ${\bm \alpha}$ and ${\bm \beta}$ are computed by minimization an energy function
\begin{equation}\label{equ:mmmeasure}
	E({\bm \alpha}, {\bm \beta}) = w_{1} E_{geoDist} + w_{2} E_{shape\_prior},
\end{equation}
where $E_{geoDist} = \sum_{i=1}^{44} ||f_i({\bm \alpha}, {\bm \beta}) - M_i||_2$ assesses the error of geodesic distance on the parameterized human body, $E_{shape\_prior}$ determines the shape prior error, with respect to mixed Gaussian distribution.
The above energy is minimized with the Particle Swarm Optimization method \cite{shi1999empirical,oikonomidis2011BMVC}.

\textbf{Parameter values.}
We set $w_1=8.0$ for $E_{geoDist}$ and $w_2=0.8$ for $E_{shape\_prior}$ in Eq.~\ref{equ:mmmeasure}.

\subsection{Pose Reconstruction}

\begin{figure}
	\centering
	\includegraphics[width=0.88\columnwidth]{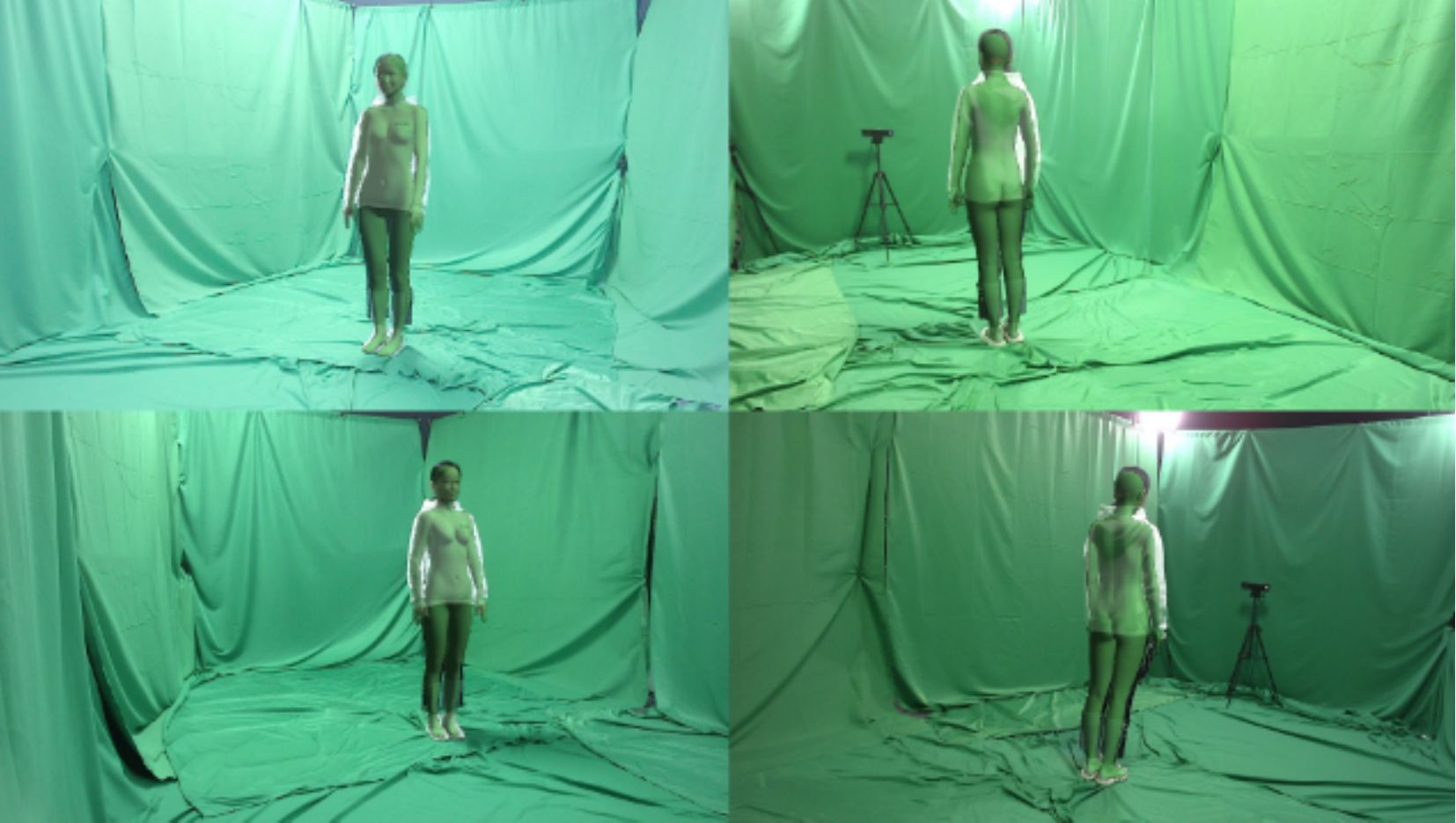}
	\caption{We reconstruct body mesh from multi-views as ground truth.}
	\label{fig:three}
\end{figure}

With four multi-view images for each frame, a 3D pose is reconstructed, according to the following steps.
\begin{enumerate}
	\item Estimate 18 joints on each image from each camera, with a 2D joint estimation method \cite{wei2016pose}.
	\item Validate the consistency of the estimated 2D joints from multi-views (see the following explanation).
	\item Segment foreground body from background automatically (green curtains setting).
	\item Solve the human motion by extending Eq.~\ref{equation:kpr} into multi-views, while dropping off the IUV constraints and the 3D constraints from Eq.~\ref{eq:data_term} (As both constraints are not available when constructing our datasets).
	\item Recover the 3D body mesh with the shape from measurements and the pose from multi-view images, shown in Fig.~\ref{fig:three}.
\end{enumerate}
In Step 2, it is very common that in a single-view image some joints could be invisible due to occlusion.
Therefore we propose a cross validation scheme based on multiple views.
For a certain joint, we choose any pair of cameras and use the two 2D estimations to build the 3D joint position.
The 3D joint is then re-projected with respect to the view-ports of the other two cameras, creating two projected joints.
Each projected joint is compared against the estimated 2D joint under the same view-port, and their Eucleadian distance is calculated.
Only if the distances in two view-ports are both below a certain threshold (18 pixels in our experiment), this set of four 2D estimations for this joint is considered to be consistent and reliable.
If four view-ports fail to reach consistency, we check if any three of them do. If that happens, the 2D joint estimation
in the three view-ports are treated as reliable. If such three view-ports can not be found, we have to ask for help from the previous frame.
The 3D joint from the previous frame is compared with the reconstructed 3D joint from any pair of cameras, and the distances are calculated.
The minimal distance designates the pair of cameras and their estimations are treated as reliable.

\subsection{IUV Maps}

The image-to-surface correspondence (IUV map) proposed in Densepose\cite{guler2018densepose} for human body is an essential mechanism for mapping a 2D image into a 3D geometry.
We improved this idea with a more solid implementation, and it works well for persons with loose clothes.
In Densepose certain pixels are manually sampled on each human part, and their corresponding points
on the meshed surface are manually marked as well.
Annotators are asked to determine the body silhouette if it is covered by clothes, and mark around 100 points for each human.
In this situation the burden is heavy and errors prone to happen, especially when a human instance wears a large/loose skirt.

\begin{figure}
	\centering
	\includegraphics[width=0.88\columnwidth]{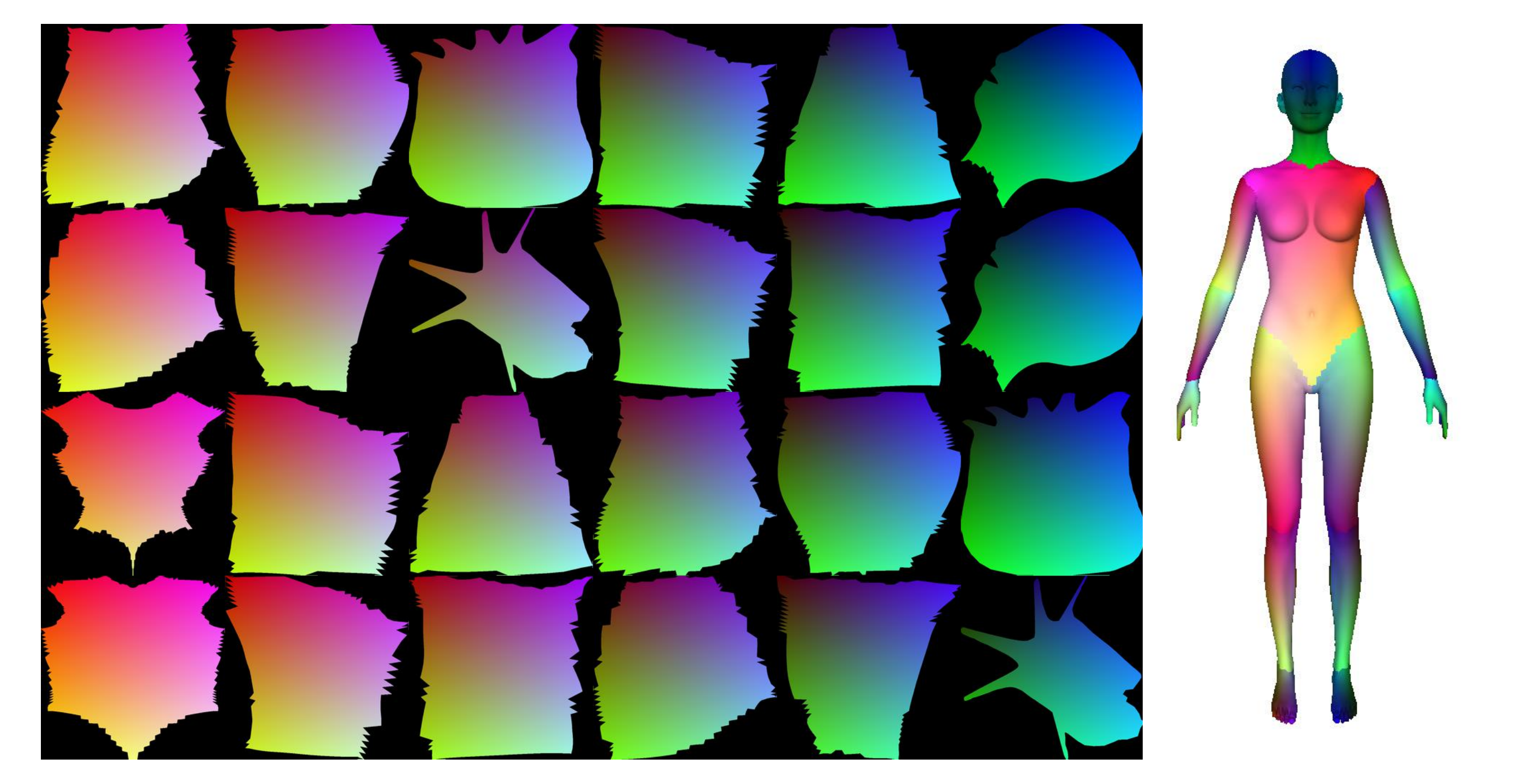}
	\caption{The IUV maps of the 24 body parts. }
	\label{fig:five}
\end{figure}

We adopt a quite different scheme for computing the part label and the $(u,v)$ coordinate for each pixel, requiring much less human intervention.
The reconstructed 3D mesh by measurements is re-projected according to the view-ports of the cameras, creating four images.
As each mesh vertex has a $(u,v)$ coordinate and a part label, it is trivial to compute $(u,v)$ and label for each pixel of these images.
Due to the topological complexity of human meshes, we follow Densepose and segment a human mesh into 24 parts, and define a $uv$-field on each part,
as shown in Fig.~\ref{fig:five}.

%% file: 5-Results.tex
\section{Results and Evaluations}

We demonstrate the power and effectiveness of our system, by reconstructing 3D human bodies from both live streams and in-the-wild videos of various scenes  (\S~\ref{subsec:test-on-real-data}).
We quantitatively compare the accuracy of our results with state-of-the-art 3D pose and/or mesh reconstruction methods (\S~\ref{sub:quatitative}). Our method is also qualitatively compared against four most related state-of-the-art methods (\S~\ref{subsec:comparisons}).  
In \S~\ref{section:ablation}, we evaluate different part the key components of our system by dropping off each term at one time for both multi-task network and optimization procedure. Our results are best seen in the accompanying video.


\begin{center}
	\begin{table}
		\centering
		\begin{tabular}{cc}
			\hline
			Component& Time (ms)\\
			\hline
			Human detection (YOLO)& 8.554\\
			Image Preprocess& 0.312\\
			Multi-task Prediction& 18.269\\
			Pose Reconstruction& 10.232\\
			Others& 11.678\\
			\hline
			Shape Recon (only for $1$st frame)& 328.227\\
			\hline
		\end{tabular}
		\caption{Running time of each component in our system.}
		\label{table:timing_time}
	\end{table}
	
\end{center}

\textbf{Computational time.}
Our system runs at 20 fps on a desktop computer for the current implementation.
Table~\ref{table:timing_time} reports the detailed timing for each component in our processing pipeline.
All execution time is collected on a computer with an Intel i7 CPU and a nvidia Geforce GTX 2080Ti GPU.
Apart from the shape reconstruction, which is done only once in the first frame, the total processing time for one cycle is under 50 ms.

\subsection{Test on live streams and in-the-wild videos}
\label{subsec:test-on-real-data}
\begin{figure}
	\centering
	\includegraphics[width=0.78\columnwidth]{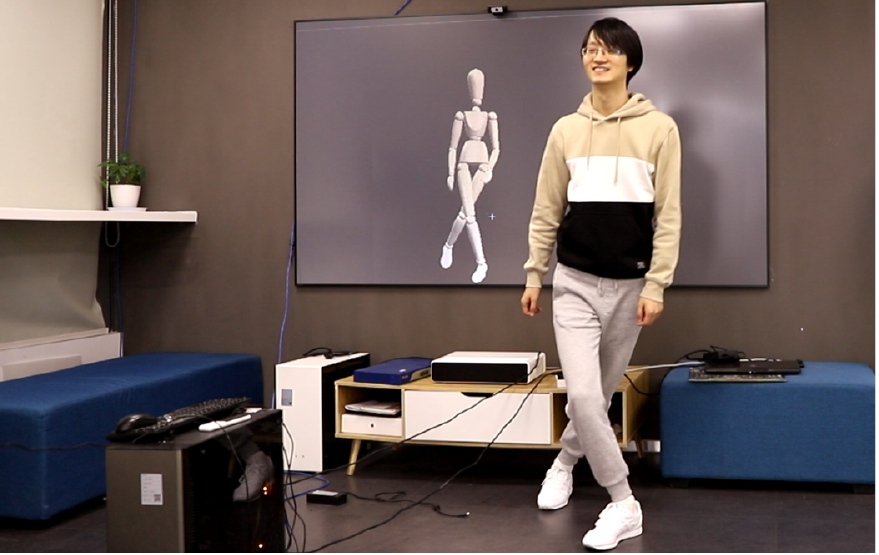}
	\caption{The reconstructed 3D poses from our system can be retargeted to animate a character in realtime.}
	\label{fig:real_scene}
\end{figure}

Our system reconstructs 3D human poses and full-body geometry models from single images in realtime, and
the reconstructed 3D poses is retargeted to animate a character in realtime (See Fig.~\ref{fig:real_scene}).
Our technology has potentials in applications such as game character control, embodied VR, sport motion analysis and reconstruction of community video.

\begin{figure}[]
    \subfigure[]{
        \centering
        \begin{minipage}[c]{0.3\linewidth}
            \includegraphics[width=2.6cm]{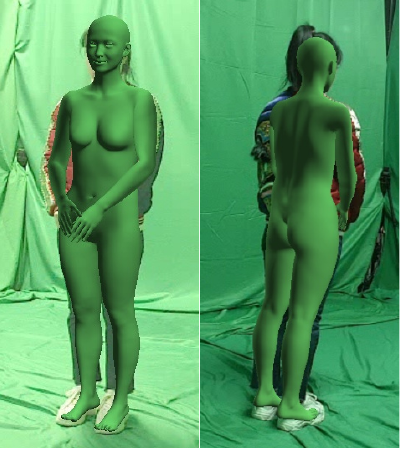}
        \end{minipage}
        }
    \subfigure[]{
        \centering
        \begin{minipage}[c]{0.3\linewidth}
            \includegraphics[width=2.6cm]{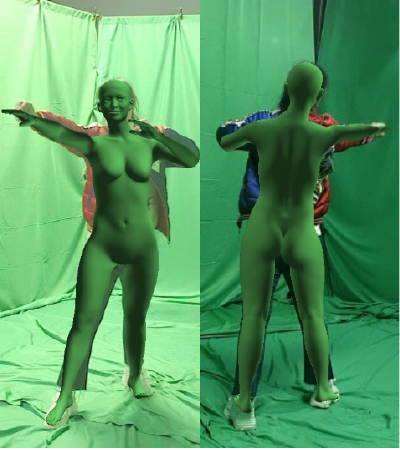}
        \end{minipage}
        }
    \subfigure[]{
        \centering
        \begin{minipage}[c]{0.3\linewidth}
            \includegraphics[width=2.6cm]{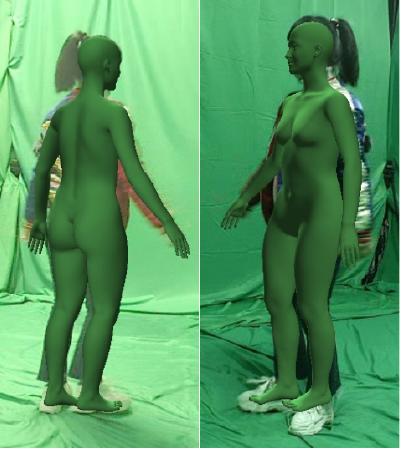}
        \end{minipage}
        }
    \caption{On the left of each subfigure is the input view from which the 3D model is reconstructed. On the right, the model is rendered and overlaid in a different view. }
    \label{fig:multi_view}
\end{figure}

\begin{figure*}
	\centering
	\includegraphics[width=1.77\columnwidth]{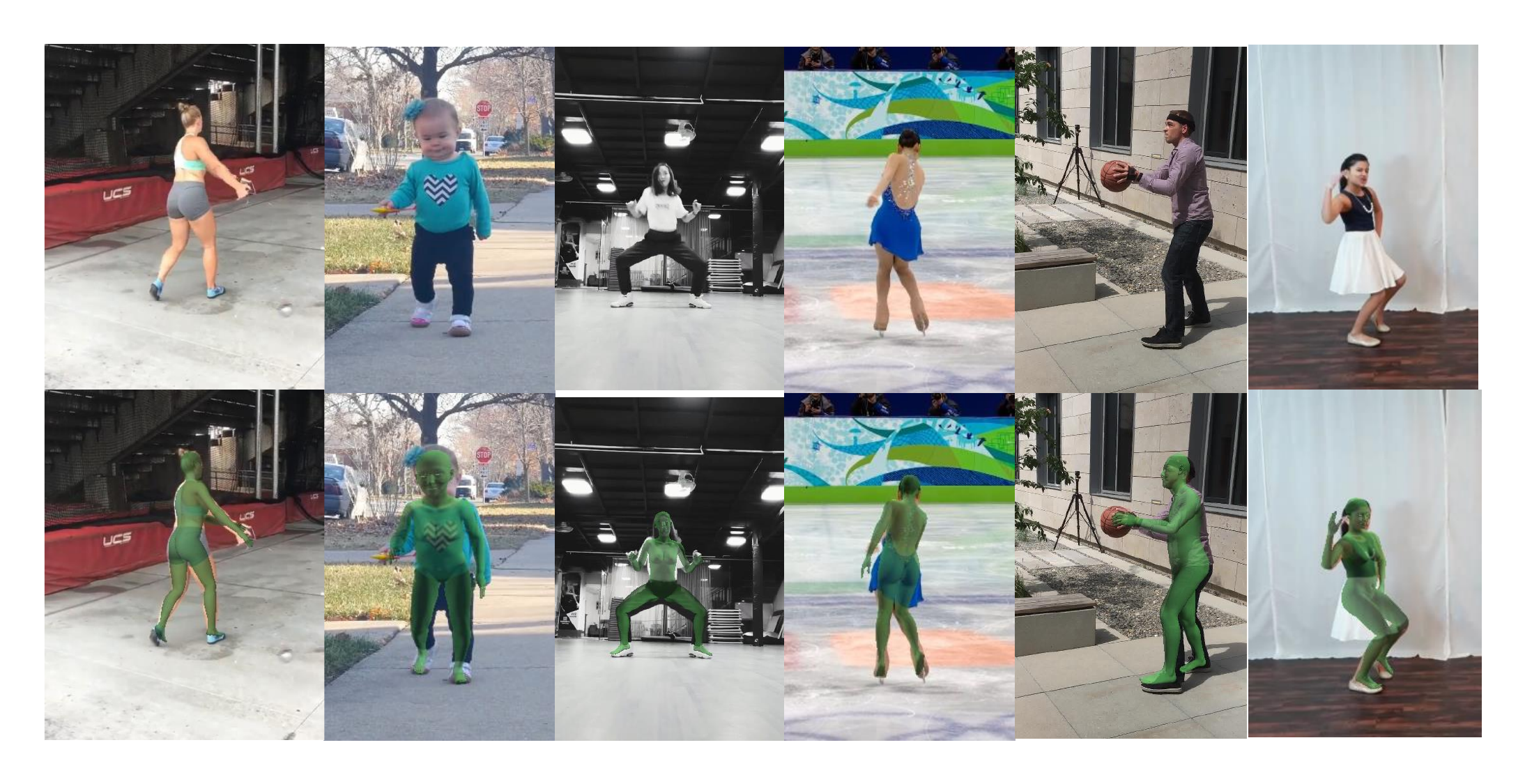}
	\caption{Our reconstructed 3D bodies with different shapes and clothes, under occlusion, lighting and background variations.}
	\label{fig:more_results}
\end{figure*}

We also test our method on various in-the-wild videos, showing its robustness to different actors with different body shapes and clothes, even under significant occlusion, lighting and background changes. Fig.~\ref{fig:more_results} shows some excerpted frames. 
Besides, we also present other-view overlay results in Fig.~\ref{fig:multi_view} by using our multi-view test dataset, which reflects the accuracy of human reconstruction achieved by our method. 
Please refer to the accompanying video for more vivid results.


\subsection{Quantitative Evaluation}
\label{sub:quatitative}
We evaluate the performance of our method on two popular test benchmarks: 3DPW \cite{Marcard_2018_ECCV} (outdoor scenes) and Human3.6M \cite{ionescu2013human3} (indoor scenes). Following the standard protocol for 3D pose estimation \cite{pavlakos2017CVPR} in Human3.6M, we use 5 subjects (S1, S5, S6, S7 and S8) for training, and the rest 2 subjects (S9 and S11) for testing. As for Human3.6M, we get ground truth parameters for training images using MoSh \cite{loper2014mosh} from the raw 3D Mocap markers like \cite{kanazawa2018CVPR}. As for 3DPW, we only use its testing dataset for evaluation as previous works \cite{kolotouros2019learning}. Note that Human3.6M and 3DPW have different skeleton configurations from ours, we therefore learn a linear regressor for a mapping which maps our mesh vertices to 17 joints defined in Human3.6M, as did in \cite{kanazawa2018CVPR}. For evaluation, we adopt averaged skeleton dimensions computed from the training set to rescale our reconstruction human, as did in \cite{pavlakos2017CVPR}.

The results are shown in Table~\ref{table:human36m2}. Our single image based method is even competitive to the video-based VIBE \cite{kocabas2020vibe} on Human3.6M and 3DPW. The results also show that our newly collected data improves the performance further, especially on 3DPW (wild), with improved wild generalization.
\begin{table}[t]
	\centering
	\footnotesize
	\setlength{\tabcolsep}{0.4mm}{
	\begin{tabular}{c|c|c|c}
		\hline
		 Method & H36M-P1 $\downarrow$ & H36M-PA$\downarrow$ & 3DPW-PA$\downarrow$ \\
		\hline
		
		Mehta {\itshape et al.}~\cite{mehta2017TOG} & 80.5 & - & - \\
		Pavlakos {\itshape et al.}~\cite{pavlakos2018CVPR} & 56.2 & 41.80 & - \\
		Sun {\itshape et al.}~\cite{sun2018ECCV} & 49.6 & 40.60 & - \\
		Zhou {\itshape et al.}~\cite{zhou2019hemlets} & \textbf{39.9} & \textbf{32.1} & - \\
		\hline
		Bogo {\itshape et al.}~\cite{bogo2016ECCV} & 82.3 & - & - \\
		Kanazawa {\itshape et al.}~\cite{kanazawa2018CVPR} & 87.97 & 56.8 & 76.7 \\
		Xiang {\itshape et al.}~\cite{xiang2019CVPR} & 58.3 & - & - \\
		kolotouros {\itshape et al.}~\cite{kolotouros2019convolutional} & 74.7 & 50.1 & 70.2 \\	
		kolotouros {\itshape et al.}~\cite{kolotouros2019learning} & - & 44.3 & 59.2 \\
		Joo {\itshape et al.}~\cite{joo2020eft} & - & 45.2 & 55.7 \\
		Kocabas {\itshape et al.}~\cite{kocabas2020vibe} & 65.6 & 41.4 & \textbf{51.9} \\
		\hline
		Ours (wild image + H36M) & 66.3 & 47.2 & 64.1 \\
		Ours (wild image + H36M + ours) & 63.7 & 41.8 & 53.2 \\
		\hline
	\end{tabular}}
	\caption{Quantitative Evaluation on Human3.6M (indoor) and 3DPW (outdoor). The number of H36M-P1 is the Mean Per Joint Position Error (MPJPE) in millimeter on Human3.6M, while the number of H36m-PA is the MPJPE on Human3.6M after procrustes alignment (PA). And 3DPW-PA is the MPJPE on 3DPW after PA. Our method achieves competitive performance against previous works.}
	\label{table:human36m2}
\end{table}


\begin{table}[t]
	\centering
	\footnotesize
	\begin{tabular}{c|c|c}
		\hline
		Methods & PCKh@0.5 $\uparrow$ & MPJPE-P1 $\downarrow$  \\ 
		\hline
		2D & 96.3 & - \\
		2D + IUVs & 97.5 & - \\
		2D + POFs & 96.2 & 51.8 \\
		2D + IUVs + POFs & \textbf{97.8} & \textbf{49.8} \\
		\hline
	\end{tabular}
	\caption{Ablation experiments on the effect of multi-task learning. In the experiment, we modify Fig.~\ref{fig:network} to set different task combinations. Our results shows that IUVs is beneficial to 2D joint detection and part orientation predictions.}
	\label{table:ablastion_multitask}
\end{table}

\begin{table}[t]
	\centering
	\footnotesize
	\setlength{\tabcolsep}{0.4mm}{
		\begin{tabular}{c|c|c|c}
			\hline
			Methods & H36M-P1$\downarrow$ & H36M-PA$\downarrow$ & 3DPW-PA$\downarrow$ \\
			\hline
			2D & 121.3 & 103.4 & 116.8 \\
			2D + mask & 118.3 & 105.1 & 110.9 \\
			2D + mask + 3D & 64.7 & 43.4 & 51.8 \\ 
			2D + mask + 3D + IUV & \textbf{63.7} & \textbf{41.8} & \textbf{51.1} \\
			2D + mask + 3D + IUV + temporal & 65.9 & 42.4 & 53.5 \\
			\hline
	\end{tabular}}
	\caption{Ablation study on the importance of each energy term in optimization. We report the MPJPE/MPJPE-PA on Human3.6M (indoor) and 3DPW (outdoor) on 5 experiments. All five experiments share the same network outputs but differ in the energy terms in optimization.}
	\label{table:ablastion_components}
\end{table}

\subsection{Comparisons with state-of-the-art methods}
\label{subsec:comparisons}
\begin{figure}
    \centering
    \subfigure[Comparison our method with SPIN \cite{kolotouros2019learning}]{
        \begin{minipage}[b]{0.8\linewidth}
            \includegraphics[width=1\linewidth]{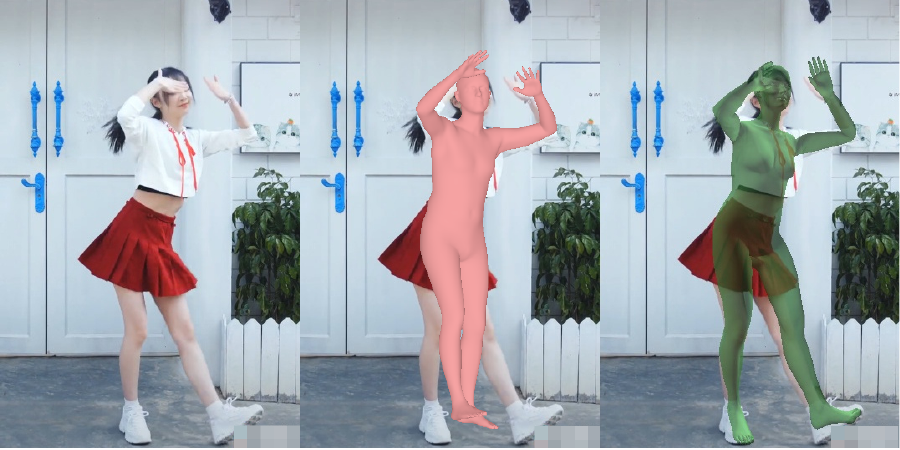}
    \end{minipage}}
    \subfigure[Comparison our method with VIBE \cite{kocabas2020vibe}]{
        \begin{minipage}[b]{0.8\linewidth}
            \includegraphics[width=1\linewidth]{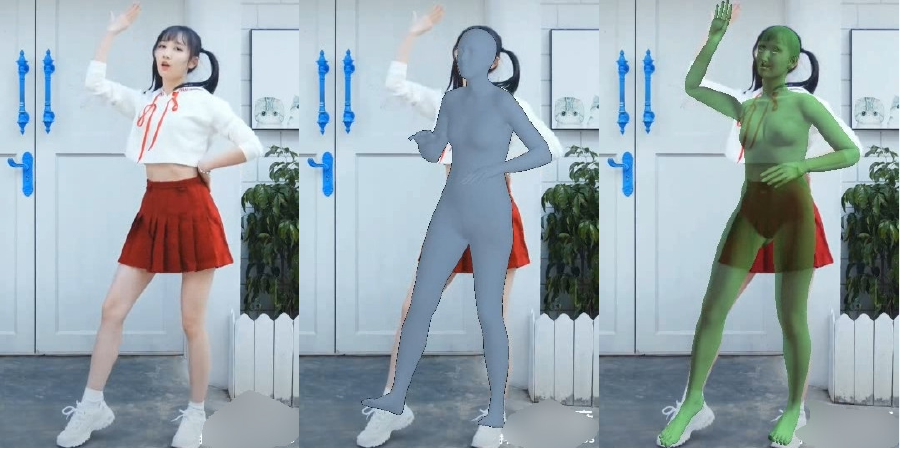}
    \end{minipage}}
\caption{From left to right: the input, SPIN/VIBE and our results. Our method produces better image-model alignments than SPIN and VIBE.} 
\label{fig:comparison_spin}
\end{figure}

To show the efficiency of our method, we compare against two state-of-the-art regression-based methods, one is single frame method, SPIN \cite{kolotouros2019learning}, the other is video-based method, VIBE \cite{kocabas2020vibe}. Fig.~\ref{fig:comparison_spin} shows the result of a side-by-side comparison. It is obvious that our method achieves better image-model alignments than SPIN and VIBE. 
Regression-based methods usually achieves global image-model alignments quickly, but at the cost of low quality. This type of nonlinear prediction is also uneasy to control, due to the mutual effect between human pose and shape. We decouple them, and since 2D joint positions and dense image-to-surface correspondence (IUV map) offer better image-model alignments, and 3D part orientation helps to avoid depth ambiguity, our method produces more accurate body reconstruction than SPIN and VIBE.

Furthermore, SPIN does not guarantee temporal stability because it regresses different body shapes from different images in a sequence, while VIBE fixes it, but it is not in realtime.
Please refer to Fig.~\ref{fig:comparison_spin} and the accompanying video for the comparison results.



\begin{figure}[h]
    \centering
    \subfigure[]{
        \begin{minipage}[t]{0.15\linewidth}
            \centering
            \includegraphics[width=1.5cm]{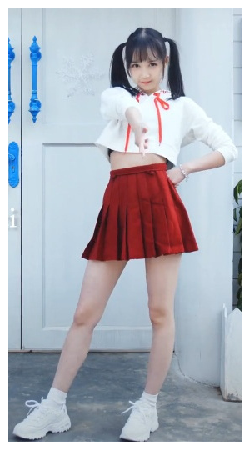}
        \end{minipage}
        }\hspace{3mm}
    \subfigure[]{
        \begin{minipage}[t]{0.30\linewidth}
            \centering
            \includegraphics[width=3.0cm]{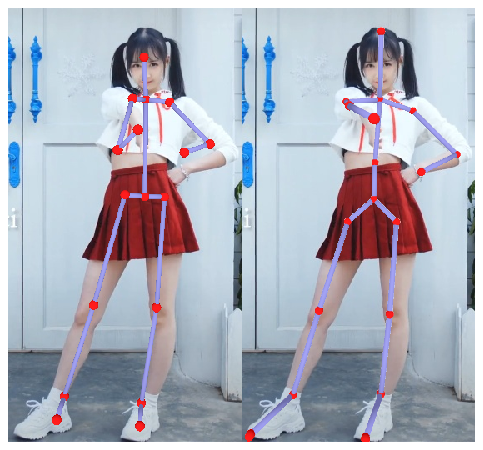}
        \end{minipage}
        }\hspace{3mm}
    \subfigure[]{
        \begin{minipage}[t]{0.36\linewidth}
            \centering
            \includegraphics[width=3.0cm]{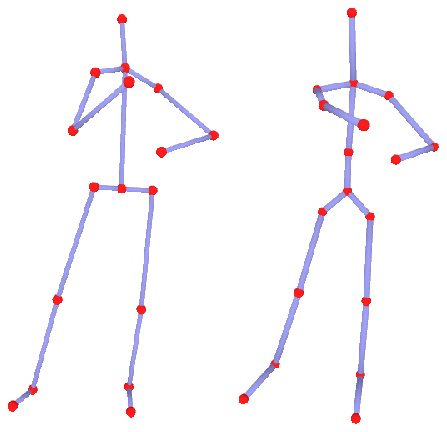}
        \end{minipage}
        }\hspace{1mm}
    \caption{Comparison of our results with VNect. (a) the input image; (b) the final results of VNect (left) and our method (right); (c) the raw network predictions of VNect (left) and our method (right).}
    \label{fig:comparison_vnect}
\end{figure}

Our method is also compared against a recent realtime system, VNect \cite{mehta2017TOG}, though it captures poses only.
We compare not only the raw network outputs, but also the final fitting results (see Fig.~\ref{fig:comparison_vnect}). It is obvious that our method achieves better pose reconstruction than VNect. Two reasons account for this. Firstly, our multi-task network produces more accurate 3D joint positions, as shown in Fig.~\ref{fig:comparison_vnect}(c). 
Secondly, VNect initializes bone lengths for forearm and upper arm to an improper ratio, as seen in Fig.~\ref{fig:comparison_vnect}(b), while our initialization matches with person in image very well. VNect initializes skeleton by averaging 3D joint positions from the CNN output at the beginning, which is thus very sensitive to single CNN outputs (3D joint positions). We instead utilize more image features, including 2D joints, 3D part orientation and IUV maps, to reconstruct human bodies more accurately and robustly.
Please refer to the accompanying video for more comparison results.



\begin{figure}
	\centering
	\includegraphics[width=0.9\columnwidth]{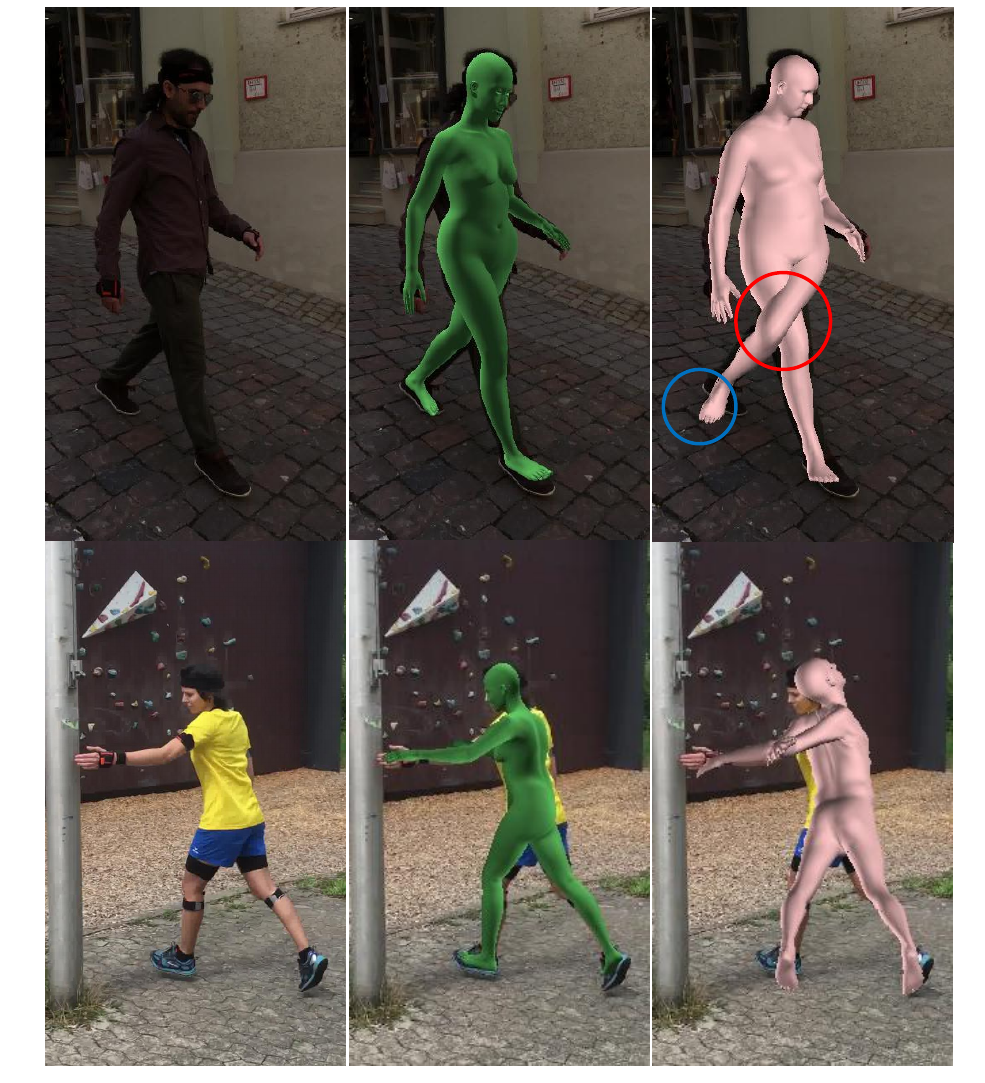}
	\caption{Comparison of our method (middle) with SMPLify (right). }
	\label{fig:smplify_vs_ours}
\end{figure}

SMPLify \cite{bogo2016ECCV} is a somewhat hybrid method: it fits the SMPL model by optimizing regressed 2D joints without user intervention.
Fig.~\ref{fig:smplify_vs_ours} shows the results given by SMPLify and our method.
At least three issues about SMPLify can be interpreted from the figure.
First, SMPLify is more vulnerable to depth ambiguity than our method, as can be seen from images in the first row.
This is because SMPLify relies on 2D joint reprojection alone for model fitting, and this is insufficient to robustly establish a 3D pose.
Second, there is no constraints for foot orientation in SMPLify.
Last, it is easy for SMPLify to fall into a local minima, as shown in second row.
Our method has hardly convergence problem. For a single image, the initial shape from the average of our human model. We use it and predicted 2D joint positions and 3D bone directions to roughly estimate the root position and joint angles as initial pose. The optimization problem is over-constrained. 

\subsection{Ablation Study}
\label{section:ablation}

We have designed a realtime multi-task network that regresses more features than any other deep learning based methods.
We believe that more features offer more visual cues that facilitate pose and shape reconstruction.
In this section we justify this belief with experiments, evaluating the role of each regressed features in both the CNN regression and the body reconstruction process.


\textbf{Quantitative analysis for multi-task network}
To evaluate the importance of our multi-task design, we modify the network in Fig.~\ref{fig:network} into 4 structures with different configurations: a) 2D joint detection only, b) 2D joint detection + IUV branch, c) 2D joint detection + POFs, d) 2D joint detection + IUV + POFs.
All these networks are trained with the same training dataset, and tested on our validation dataset, which contains 11 different subjects not in the training dataset. 
For metrics of 2D joint positions, we report PCKh@0.5 \cite{newell2016ECCV} (the higher the better). For 3D part orientation, we scale the predicted 3D part orientation by the ground-truth limb length to obtain the 3D joint positions, then align the root joint position and compute the MPJPE (the lower the better). 
The results are reported in Table~\ref{table:ablastion_multitask}, which shows the power of mutual promotion of multi-task. 
We found that IUV information improves the accuracy of 3D POF orientation. The reason is that IUV maps provide the part occlusion relationship which conveys some 3D information. IUV maps are usually more abstract and more powerful than 2D landmarks in representing a human, and it is used by \cite{xu2019denserac} as input. 

\begin{figure}[h]
	\centering
	\includegraphics[width=0.9\columnwidth]{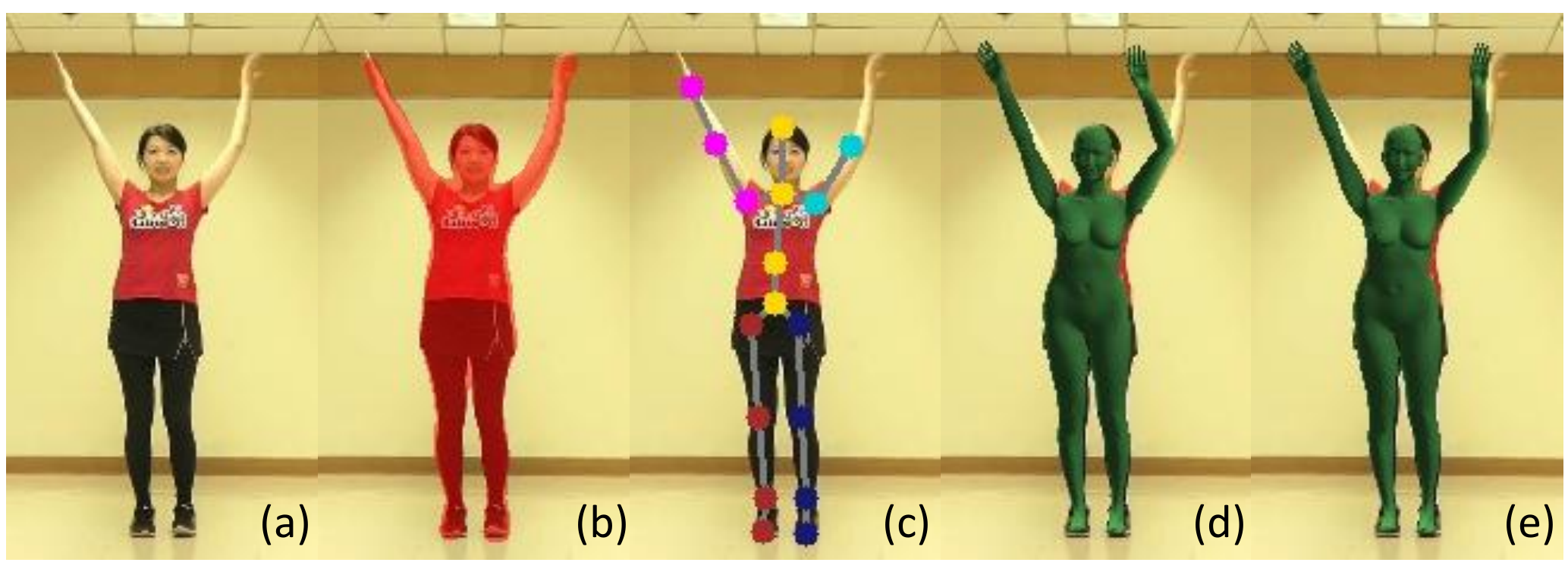}
	\caption{Importance of the mask term. Given an image (a), the network predicts a foreground segmentation mask (b) and 2D joints (c) (note the left-wrist is missing).
		If the mask is not used, the reconstructed body is problematic (d); The mask helps build a correct pose (e).}
	\label{fig:ablation_mask}
\end{figure}

\textbf{Quantitative analysis for optimization}
Table~\ref{table:ablastion_components} shows the quantitative performance, which reveals the importance of each energy term. We compare results under 5 different energy term settings: a) 2D position only; b) 2D position + mask; c) 2D position + mask + 3D part orientation; d) 2D position + mask + 3D part orientation + IUV; e) 2D position + mask + 3D part orientation + IUV + temporal. We report MPJPE on Human3.6M and 3DPW. The results shows that every energy term in our optimization is beneficial for human pose reconstruction. 
The result with temporal term shows higher errors, because it ensures temporal smooth rather than the consistency with the network output cues.
\textbf{The mask term in optimization.}
We evaluate the importance of the foreground segmentation mask by comparing the reconstructed bodies with and without this term.  Fig.~\ref{fig:ablation_mask} clearly shows the role of the mask when predicted 2D joints and 3D orientation are inaccurate, especially when some joints are missing.

\begin{figure}[h]
	\centering
	\includegraphics[width=0.8\columnwidth]{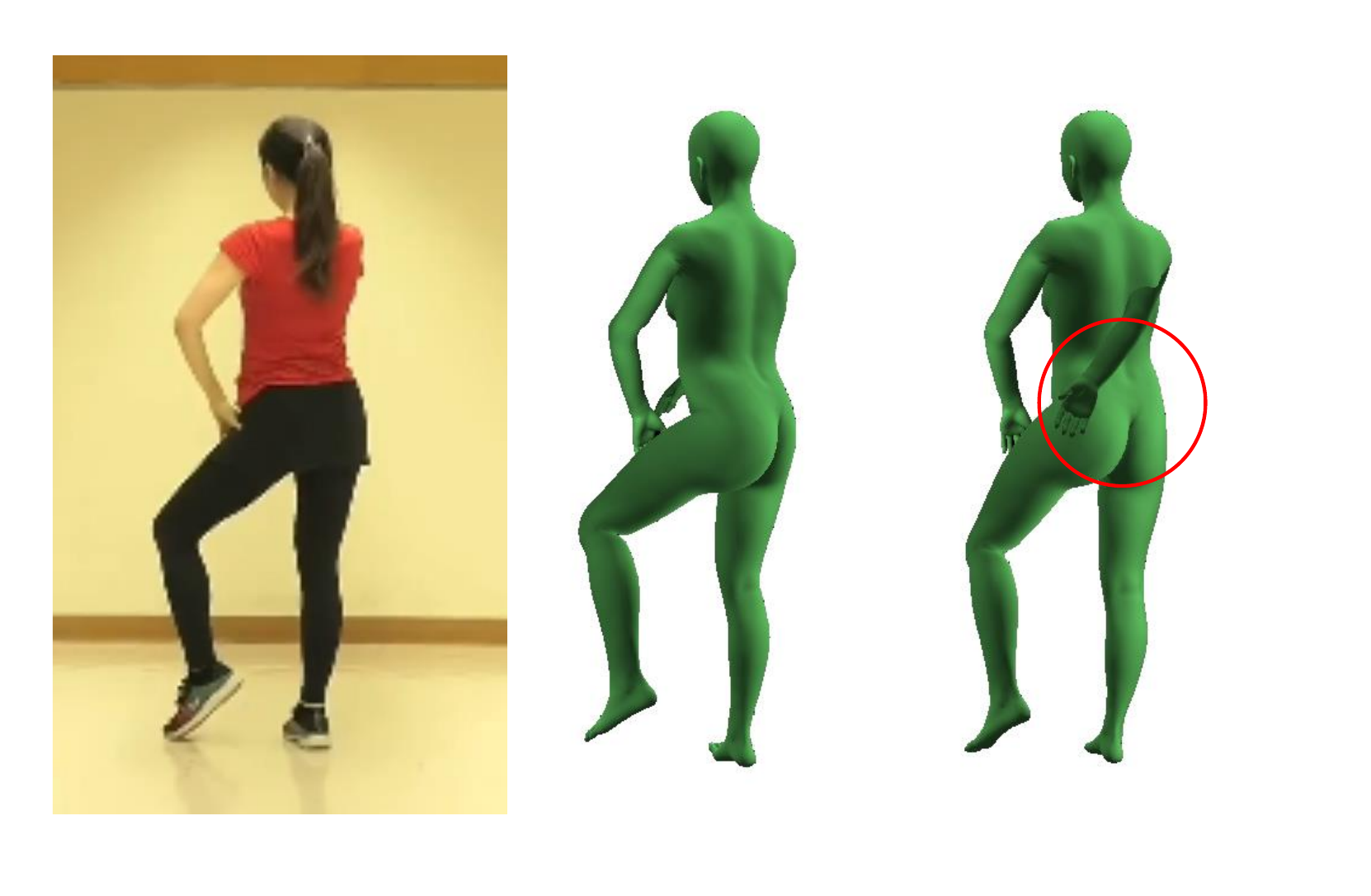}
	\caption{Importance of the 3D part orientation term. (left) input image; (middle) result with 3D part orientation term; (right) result without 3D part orientation term. }
	\label{fig:ablation_3d}
\end{figure}

\textbf{The 3D part Orientation term in optimization.}
Fig.~\ref{fig:ablation_3d} shows an example with and without the 3D orientation term. The use of the 3D orientation term significantly reduces the reconstruction ambiguity of 3D poses.

\begin{figure}
    \centering
    \subfigure[Shape reconstructed with the IUV term (middle) manifests the body weight better than that without the IUV term (right).]{
        \begin{minipage}[b]{0.9\linewidth}
            \includegraphics[width=1\linewidth]{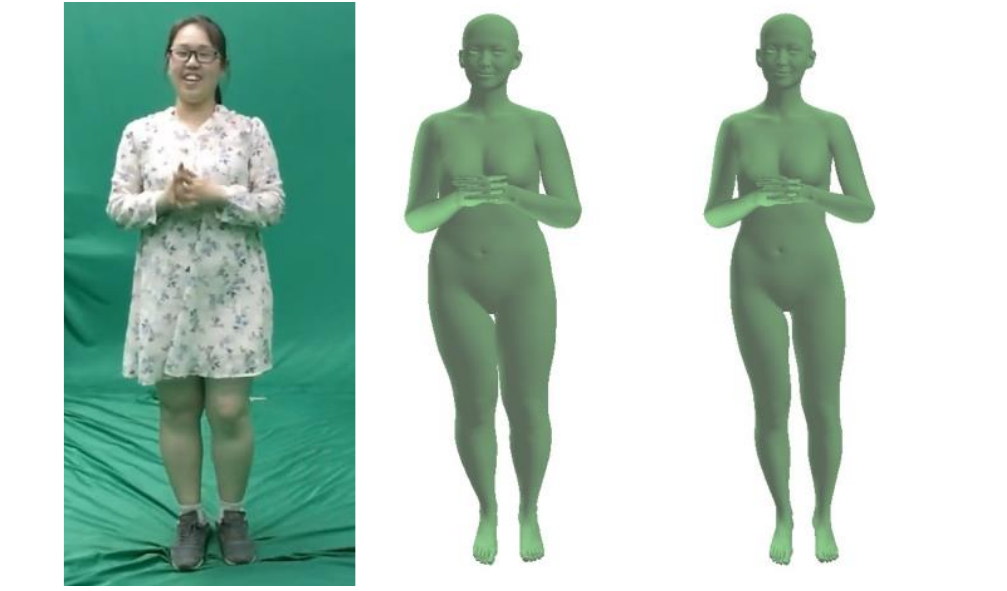}
    \end{minipage}}
    \subfigure[Pose reconstructed with (middle) and without (right) the IUV term.]{
        \begin{minipage}[b]{0.9\linewidth}
            \includegraphics[width=1\linewidth]{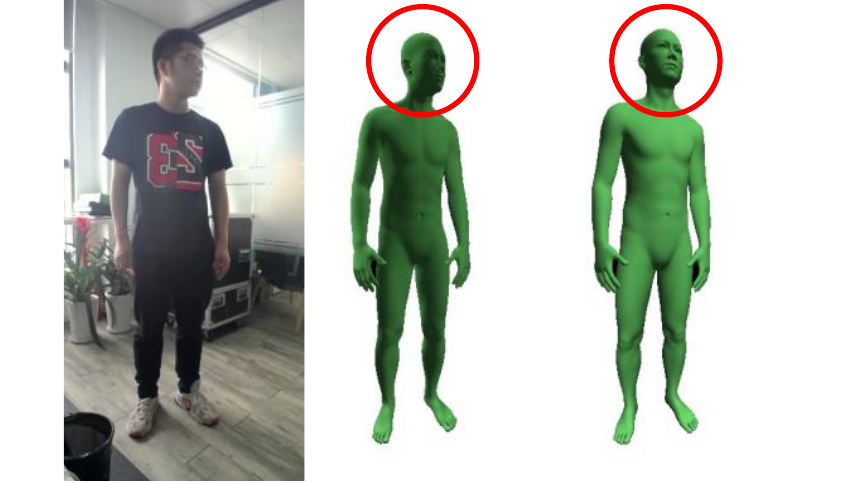}
    \end{minipage}}
\caption{The importance of IUV term for shape and pose reconstruction.} 
\label{fig:ablation_iuv}
\end{figure}

\textbf{The IUV term in optimization.}
An IUV map plays important roles in both 3D body geometry reconstruction and pose estimation. We evaluate the importance of IUV by dropping off this term in shape reconstruction and pose estimation, respectively.
Fig.~\ref{fig:ablation_iuv}(a) shows a side-by-side comparison while reconstructing an over-weighted lady. Using IUV term gives more accurate body model, because IUV terms impose detailed geometry model constraints from dense correspondences.


For pose reconstruction, Fig.~\ref{fig:ablation_iuv}(b) shows that reconstructed examples with and without IUV term. It is obvious that IUV term helps recover more accurate result, especially for body orientation. 


\subsection{Limitations}
With no exceptions, our method suffers from several limitations.
First, we observed failure cases when a significant part of the target person is either occluded by other objects or out of image boundary.
Occlusion is the biggest issue and it imposes more challenge for RGB camera than depth camera based methods.
Second, our method also fails for complicated or uncommon poses, particularly those in sports videos, such as gymnastics and skydiving.
The main reason is that such data is not adequate in training dataset.
Third, our system does not have specific hand pose detector (as did in \cite{joo2018total}) and each hand is associated with only one joint, therefore the reconstructed hands are sometimes incorrectly oriented.
Finally, our CNN does not handle multiple bodies at this moment, but can easily extended to support this.
Solving the above mentioned problems points to interesting future directions.

\section{Conclusions}
We have presented a method for reconstructing the 3D pose and shape of a human, in a stable and consistent manner, from a single RGB video stream at more than 20 Hz.
Our approach employs a multi-task CNN that regresses five human anatomical features simultaneously,
which are further cooked with a kinematic pose reconstruction and shape modeling algorithm,
producing a temporally stable 3D reconstruction of the full-body.
In contrast to most existing approaches, our approach can operate on any input image fully automatically,
without strict prescribed bounding boxes, and independent of expensive initialization.
We test and evaluate our system in a variety of challenging realtime scenarios, including live streaming from commercial cameras, as well as in community videos.
Results demonstrate that our approach compares to offline state-of-the-art monocular RGB methods qualitatively and advances the realtime 3D body reconstruction methods with a significant step.